\documentclass[lettersize,journal]{IEEEtran} 
\usepackage{color}
\usepackage[hypertexnames=false]{hyperref} 
\usepackage{url} 
\usepackage{graphicx}
\usepackage{amsmath}
\usepackage{booktabs}
\usepackage{amssymb}
\usepackage{multirow}
\usepackage{array}
\usepackage{algorithmic}
\usepackage{algorithm}
\usepackage{bm} 
\usepackage[dvipsnames,table]{xcolor}
\usepackage{amsthm}
\usepackage{threeparttable} 
\newtheorem{definition}{Definition}

\usepackage{graphicx}

\DeclareGraphicsExtensions{.pdf,.png,.jpg}

\definecolor{OliveGreen}{rgb}{0.33, 0.42, 0.18}

\ifCLASSINFOpdf 
\else
\fi
\begin{document}
\title{MoE-Enhanced Explainable Deep Manifold Transformation for Complex Data Embedding and  Visualization} 

\author{Zelin Zang\IEEEauthorrefmark{1},
        Yuhao Wang\IEEEauthorrefmark{1},
        Jinlin Wu,
        Hong Liu,
        Yue Shen,
        Zhen Lei$\IEEEauthorrefmark{2}$
        and~Stan Z. Li$\IEEEauthorrefmark{2}$%
\thanks{\IEEEauthorrefmark{1}These authors contributed equally to this work.}%
\thanks{\IEEEauthorrefmark{2}Corresponding authors: Zhen Lei (zhen.lei@ia.ac.cn) and Stan Z. Li (Stan.ZQ.Li@westlake.edu.cn).}%
\thanks{Zelin Zang is with Westlake University, Hong Kong Institute of Science \& Innovation CAS, and Tsientang Institute for Advanced Study. email: zangzelin@westlake.edu.cn}% <-this % stops a space
\thanks{
  Yuhao Wang and Stan.Z Li are with Westlake University. Hong Liu is with School of Information and Electrical Engineering, Hangzhou City University, Hangzhou, 310015 China and Academy of Edge Intelligence Hangzhou City University, Hangzhou City University, 310015 China.
  Jinlin Wu is with CAIR, HKISI-CAS; 
  State Key Laboratory of Multimodal Artiﬁcial Intelligence Systems (MAIS), CASIA.
  Zhen Lei is with CAIR, HKISI-CAS; MAIS, CASIA; and School of Artiﬁcial Intelligence, University of Chinese Academy of Sciences (UCAS). Yue Shen is with Ant Group.}}


\markboth{MoE-Enhanced Explainable Deep Manifold Transformation}%
{Zang \MakeLowercase{\textit{et al.}}: MoE-Enhanced Explainable Deep Manifold Transformation}



\IEEEpubid{\scriptsize
\textit{This article has been accepted for publication in IEEE Transactions on Pattern Analysis and Machine Intelligence. DOI: 10.1109/TPAMI.2026.3705888.} \quad
\textcopyright~2026 IEEE. Personal use of this material is permitted. Permission from IEEE must be obtained for all other uses, in any current or future media, including reprinting/republishing this material for advertising or promotional purposes, creating new collective works, for resale or redistribution to servers or lists, or reuse of any copyrighted component of this work in other works.} 

\maketitle

\begin{abstract}
  Dimensionality reduction (DR) plays a crucial role in various fields, including data engineering and visualization, by simplifying complex datasets while retaining essential information. 
  % However, the challenge of balancing DR accuracy and explainability remains crucial, particularly for users dealing with high-dimensional data. 
  However, achieving both high DR accuracy and strong explainability remains a fundamental challenge, especially for users dealing with high-dimensional data.
  Traditional DR methods often face a trade-off between precision and transparency, where optimizing for performance can lead to reduced explainability, and vice versa. This limitation is especially prominent in real-world applications such as image, tabular, and text data analysis, where both accuracy and explainability are critical. To address these challenges, this work introduces the MoE-based Explainable Deep Manifold Transformation (DMT-ME). The proposed approach combines a geometry-aware hyperbolic mapper with Mixture of Experts (MoE) models, where sparse expert specialization provides the main representational gain and the hyperbolic component offers an additional refinement for structurally complex data. DMT-ME enhances DR accuracy primarily through MoE-based sparse routing and structure-aware matching, while also improving explainability by explicitly linking input data, embedding outcomes, and key features through the MoE structure. Extensive experiments demonstrate that DMT-ME consistently achieves superior performance in both DR accuracy and model explainability, making it a robust solution for complex data analysis. The code is available at \url{https://github.com/zangzelin/code_dmtme}. 
\end{abstract}


% \fi
%
\section{Introduction}

\begin{figure}[t]
    \centering
    \includegraphics[width=0.48\textwidth]{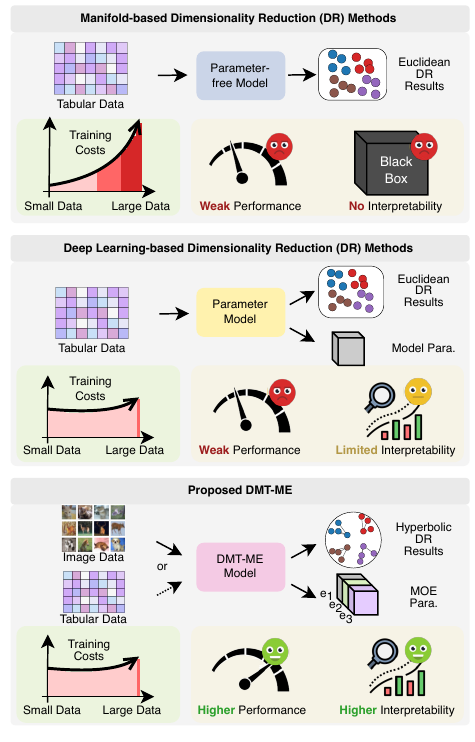}
    \vspace{-0.4cm}
    \caption{\textbf{Overview of the proposed DMT-ME network.} The figure compares three DR approaches: manifold-based, deep learning-based, and the proposed DMT-ME. DMT-ME leverages an MoE strategy to effectively process both image~\cite{zang2022dlme} and tabular~\cite{zang2023udrn} data, offering improved performance, lower training cost, and enhanced explainability across different data scales.
    }
    \vspace{-0.4cm}
    \label{fig_intro_motivation}
\end{figure}
Dimensionality reduction~(DR)~\cite{bui2023dimension,charte2021reducing} is a fundamental tool in high-dimensional data analysis. It projects data into a low-dimensional space while retaining crucial structural information, thus reducing computational cost, storage requirements, and visualization complexity~\cite{yu2024dimensionality,bera2021dimensionality}. Among its various applications, low-dimensional visualization is particularly significant, enabling intuitive pattern recognition and interpretation in high-dimensional settings~\cite{ayesha2020overview,ray2021various,hojjati2023dasvdd}.
However, DR for visualization~\cite{maaten2008visualizing} is among the most challenging and constrained forms of representation learning. Unlike general embedding learning, which may prioritize performance or compactness, visualization demands both high accuracy and strong explainability~\cite{kabir2023performance,imrie2023multiple}. Embeddings must preserve meaningful structures at multiple levels (e.g., local neighborhoods and global hierarchies)~\cite{piaggesi2024dine,bharadiya2023tutorial,duque2022geometry} while remaining visually interpretable---highlighting a critical trade-off between optimization accuracy and cognitive clarity.

DR methods typically fall into two main categories: parameter-free, manifolds-based techniques~\cite{hajderanj2024novel,you2024gnumap,yi2024structure} and deep learning-based approaches~\cite{luo2020dimensionality}. Manifold-based methods, such as tSNE~\cite{van2008visualizing} and UMAP~\cite{UMAP}, offer speed and adaptability for small datasets~\cite{ayesha2020overview}, using nonlinear mappings to reveal latent structures. Deep learning-based methods, including parametric UMAP~\cite{sainburg2021parametric,xu2023robust} and DMT-EV~\cite{zang2024dmt}, are better suited for handling high-dimensional and complex data by leveraging neural networks. DMT-EV, notably, demonstrates both strong performance and improved explainability by eliminating irrelevant features~\cite{marcinkevivcs2023interpretable}. These methods are more scalable and generalizable, making them increasingly central to DR research.

In terms of efficiency, performance, and explainability, manifold-based and deep learning-based methods exhibit different strengths and limitations due to their design principles~\cite{ayesha2020overview}. Parameter-free methods are efficient for smaller datasets due to their non-parametric nature~\cite{zhang2021survey}, but their performance degrades as data size increases, owing to costly neighborhood computations. Deep learning-based methods scale more effectively, benefiting from model reusability and hardware acceleration, though they incur higher training costs on smaller datasets. Performance-wise, manifold-based methods excel at capturing local structure but often struggle with complex global relationships due to their reliance on Euclidean assumptions~\cite{drnovsek2021hyperbolic}. Conversely, deep learning can capture both local and global features but requires substantial data and computational resources. In terms of explainability, manifold methods rely on local similarities, which often lack consistency across the dataset, while deep models---despite capturing richer semantics---remain difficult to interpret due to their black-box characteristics.

To address these limitations in global structure modeling and interpretability, we propose the \textbf{mixture of experts (MoE)-based explainable deep manifold transformation~(DMT-ME)}, as shown in Fig.~\ref{fig_intro_motivation}. Our approach integrates deep manifold learning with a MoE framework~\cite{zhou2022mixture,cai2024survey}.
We introduce a dynamic matching loss tailored to batch-sampled submanifolds, preserving local structure while ensuring global coherence. The MoE strategy improves performance and computational efficiency by enabling specialized expert networks to dynamically select relevant input features, thus avoiding bottlenecks from single-model architectures. In addition, we employ multiple Gumbel matchers to provide the MoE with orthogonal feature representations. This not only encourages diversity in expert learning but also enhances explainability. By analyzing the feature selection patterns and specialization of experts, we can infer the significance and relevance of input features. Collectively, these innovations enable DMT-ME to achieve superior performance, efficiency, and interpretability for DR-based visualization of complex datasets.

The main contributions of this work are as follows:
\begin{itemize}
    \item A submanifold-based dynamic matching loss function that enhances DR accuracy by capturing global structures more effectively.
    \item A MoE strategy that strengthens explainability and stability by linking input features, embeddings, and key components explicitly.
    \item Extensive evaluations on global and local performance, time efficiency, and additional metrics to demonstrate the advantages of DMT-ME.
\end{itemize}

\section{Related Work}
\textbf{Dimension Reduction and Visualization.}
\label{sec_related_work}
DR methods can be broadly categorized into parameter-free and parametric approaches. Parameter-free methods, such as MDS~\cite{kruskal1964nonmetric}, ISOMAP~\cite{tenenbaum_global_2000}, LLE~\cite{roweis_nonlinear_2000}, and tSNE~\cite{maaten_visualizing_2008}, directly optimize the output by preserving pairwise distances, with a strong emphasis on capturing local manifold structures. However, these methods often lack generalizability and explainability, as they do not model the underlying feature relationships and are unable to handle unseen data effectively. By contrast, parametric methods, such as topological autoencoders~\cite{moor2020topological} and PUMAP~\cite{sainburg_parametric_2021}, leverage neural networks to learn continuous mappings constrained by the input space, thereby improving both generalizability and interpretability. NeuroDAVIS~\cite{maitra_neurodavis_2024} advances DR performance by jointly preserving local and global data structures. Despite these improvements, many existing methods still face challenges in designing effective loss functions and training strategies, particularly when dealing with complex biological data. Integrative approaches, such as causal representation learning, exemplified by Cells2Vec~\cite{rajwade_cells2vec_2023}, represent a promising direction for enhancing both the explainability and robustness of DR methods.

\textbf{Interpretability and Explainability of DR Methods.}
Interpretability refers to understanding a model's internal mechanisms, while explainability focuses on how easily the model's outputs can be understood by users~\cite{zhang2021survey,imrie2023multiple}. In the context of DR, explainability-enhancing methods, such as DMT-EV~\cite{zang2024dmt}, employ saliency maps and interactive visualization tools to highlight how input features influence the low-dimensional embeddings. Other techniques, including DimReader~\cite{RebeccaFaust2019DimReaderAL} and DataContextMap~\cite{ShenghuiCheng2016TheDC}, visually associate features with their influence in the reduced space, thereby improving user understanding of the embeddings. 
Explainability is particularly vital in domains, such as healthcare and finance, where transparent decision-making is essential~\cite{chefer_transformer_2021}. Hybrid approaches, such as DT-SNE~\cite{bibal_dt-sne_2023}, which integrate tSNE with decision tree logic, provide clearer interpretability by mapping embedding behavior to comprehensible rules, further advancing DR methods in high-stakes applications.

\textbf{Hyperbolic Embeddings.}
Hyperbolic embeddings have attracted increasing attention for their ability to model hierarchical structures more effectively than Euclidean embeddings, enabling the simultaneous preservation of local and global relationships~\cite{nickel2017poincare}. These embeddings are particularly useful in tasks such as zero-shot learning~\cite{khrulkov2020hyperbolic}, graph analysis, and hierarchical data visualization.
In the field of bioinformatics, for instance, scDHMap~\cite{tian_complex_2023} uses hyperbolic space to better capture complex cell differentiation trajectories in single-cell RNA-seq data. Likewise, D-Mercator~\cite{jankowski_d-mercator_2023} and hyperbolic informed embedding~\cite{yang_hyperbolic_2023} have extended the utility of hyperbolic embeddings to network visualization and hierarchical representation learning. Despite their advantages, hyperbolic models face challenges with numerical stability and optimization~\cite{mishne_numerical_2023}, highlighting an important direction for further research to improve scalability and robustness across applications.
A more detailed review of related work is omitted here for brevity.

\section{Problem Definition and Preliminaries}

\subsection{Data Description and Augmentation} 
This section formally describes the preprocessing, data augmentation \cite{shorten2019survey}, DR, and explainability mechanisms adopted in the proposed method. These elements are foundational for understanding the subsequent sections.
Let the input data $\mathbf{X}$ represent an image, tabular, or sequential data type. Assuming a dataset of $N$ samples, each sample is denoted as $\mathbf{x}_i$, with the dataset expressed as:
\begin{equation}
 \mathbf{X} := \{\mathbf{x}_1, \mathbf{x}_2, \ldots, \mathbf{x}_i, \ldots, \mathbf{x}_N\}.
\end{equation}
To improve model robustness and increase training sample diversity, we perform data augmentation. The augmented dataset $\mathbf{X}^{+}$ is defined as:
\begin{equation}
 \begin{aligned}
 \mathbf{X}^{+} &:= \{\mathbf{x}_1^{+}, \mathbf{x}_2^{+}, \ldots, \mathbf{x}_i^{+}, \ldots, \mathbf{x}_N^{+}\},\\
 \mathbf{x}_i^{+} &:= \tau(\mathbf{x}_i),
 \end{aligned}
\end{equation}
where each $\mathbf{x}_i^{+}$ is generated through specific augmentation techniques based on the data type.

\textbf{Tabular Data.} For tabular inputs, features are normalized to zero mean and unit variance to facilitate stable model training and uniform feature scaling \cite{zhong2020random}. Data augmentation is further achieved through neighbor-based interpolation. Specifically, each point $\mathbf{x}_i$ is augmented as follows: 
\begin{equation}
 \begin{aligned}
 \tau^\text{tab}(\mathbf{x}_i) := \lambda \mathbf{x}_i + (1 - \lambda) \mathbf{x}_i', 
 \\
 \mathbf{x}_i' \in \mathcal{N}^\text{k}(\mathbf{x}_i), \quad \lambda \sim \mathcal{U}(0, 1),
 \end{aligned}
\end{equation}\label{eq:tab_aug}
where $\lambda$ is a random interpolation coefficient drawn from a uniform distribution over the interval 0 and 1; $\mathbf{x}_i'$ is a randomly selected neighbor of $\mathbf{x}_i$; $\mathcal{N}^\text{k}(\mathbf{x}_i)$ denotes the function that retrieves the $k$ nearest neighbors of $\mathbf{x}_i$; and $\text{k}$ is a hyperparameter. 
 
\textbf{Image Data.} For image datasets, we utilize standard data augmentation techniques, such as random cropping, flipping, and rotation, to enhance the diversity and robustness of the training set
~\cite{krizhevsky2012imagenet,simonyan2014very,shorten2019survey}. 
Specifically, for a given image $\mathbf{x}_i$, we define a transformation set $\tau^\text{img}$, applied as follows:
\begin{equation}
\mathbf{x}_i^{+} = \tau^\text{img}(\mathbf{x}_i),
\end{equation}
where $\tau^\text{img}$ encompasses techniques including color jittering~\cite{zhan2020online}, random cropping~\cite{cheng2020random}, applying Gaussian blur~\cite{flusser2015recognition}, Mixup~\cite{liu2021automix}, and additional domain-specific augmentations~\cite{zangdiffaug}.

\subsection{Explanation in DR}

\begin{definition}[Additive Explanation] \label{def_additive}
Consider a scalar-valued prediction function \( f : \mathbb{R}^d \to \mathbb{R} \).
An {additive explanation} is defined as a set of component functions \( \{c_k(\mathbf{x})\}_{k=1}^{K} \), with each \( c_k : \mathbb{R}^{d_k} \to \mathbb{R} \), that satisfy the following conditions for every input \( \mathbf{x} \in \mathbb{R}^d \):
\textbf{(C1) Completeness / Faithfulness.} 
The output of the model is fully decomposed into a sum of interpretable components:
$ f(\mathbf{x}) = \sum_{k=1}^{K} c_k(\mathbf{x}).$
\textbf{(C2) Local Dependence.} 
Each component relies only on a specific transformation of the input:
$ 
c_k(\mathbf{x}) = c_k(\mathbf{z}_k(\mathbf{x})), \quad \text{where} \quad \mathbf{z}_k(\mathbf{x}) = \mathbf{x}_{\mathcal{S}_k},\; \mathcal{S}_k \subseteq [d].
$
This formulation encompasses a broad class of explanatory approaches.
\end{definition}

\vspace{0.5em}
Although this definition pertains to scalar outputs, many real-world models produce low-dimensional vector representations. This is particularly prevalent in DR settings, where the mapping \( f : \mathbb{R}^D \to \mathbb{R}^d \) transforms a high-dimensional input \( \mathbf{x} \in \mathbb{R}^D \) into a low-dimensional embedding \( \mathbf{h} = f(\mathbf{x}) \). In such cases, the purpose of explanation evolves from attributing scalar outputs to understanding the influence of individual features or feature subsets on the structure and geometry of the resulting embedding space. We formalize this notion below:

\begin{definition}[Feature Explanation in DR]
Let \( f : \mathbb{R}^D \to \mathbb{R}^d \) be the DR function and \( \{\mathbf{x}_i\}_{i=1}^N \) denote a dataset. A \textit{{feature explanation}} is a function that associates the input samples and their corresponding embeddings \( \{\mathbf{h}_i = f(\mathbf{x}_i)\} \) with a vector indicating feature-wise importance:
\begin{equation}
I_{\mathrm{feat}} : \left( \{\mathbf{x}_i\}, \{\mathbf{h}_i\}, f \right) \longrightarrow \mathcal{I}_{\mathrm{feat}} \in \mathbb{R}^{D},
\end{equation}
where the \( j \)-th element, \( \mathcal{I}_{\mathrm{feat}}[j] \), reflects the mean influence of the \( j \)-th input feature on the overall structure of the embedding space learned by the model.
\end{definition}

\begin{definition}[Group Explanation in DR]
Building on the previous concept, a \textit{group explanation} targets a collection of predefined feature subsets \( \{\mathcal{S}_1, \dots, \mathcal{S}_G\} \subseteq [D] \), assigning an importance value to each group as a whole:
\begin{equation}
I_{\mathrm{group}} : \left( \{\mathbf{x}_i\}, \{\mathbf{h}_i\}, f \right) \longrightarrow \mathcal{I}_{\mathrm{group}} \in \mathbb{R}^{G},
\end{equation}
where \( \mathcal{I}_{\mathrm{group}}[g] \) indicates the impact of feature group \( \mathcal{S}_g \) on the overall structure of the embedding space. This facilitates interpretation at a higher, and frequently more semantically relevant, level.
\end{definition}

\subsection{Gumbel-Softmax for Differentiable Selection}
The Gumbel-Softmax method~\cite{jang2016categorical} permits differentiable sampling from a categorical distribution, making it suitable for discrete selection tasks within neural networks, such as selecting feature subsets. Consider a set of logits denoted by $\mathbf{L} = (\mathbf{L}_1, \mathbf{L}_2, \ldots, \mathbf{L}_\text{D})$, where $\text{D}$ represents the number of candidates (e.g., features). The Gumbel-Softmax distribution injects Gumbel noise $\mathbf{g}_i$ and subsequently applies the softmax function,
\begin{equation}
 \mathbf{y}_i := \frac{\exp((\mathbf{L}_i + \mathbf{g}_i) / \tau)}{\sum_{j=1}^{D} \exp((\mathbf{L}_j + \mathbf{g}_j) / \tau)}
\end{equation}
where each $\mathbf{g}_i$ is independently sampled from a Gumbel(0,1) distribution, and $\tau$ serves as a temperature parameter that regulates the smoothness of the distribution. As $\tau \to 0$, the distribution converges to a categorical distribution, resulting in discrete hard selections. Conversely, larger values of $\tau$ produce softer outputs that facilitate exploration. This mechanism enables the model to learn discrete structures (like assigning a feature to a specific expert) while maintaining differentiability for gradient-based training via the reparameterization trick.

\begin{figure*} 
 \centering
 \vspace{-0.4cm}
 \includegraphics[width=0.99\textwidth]{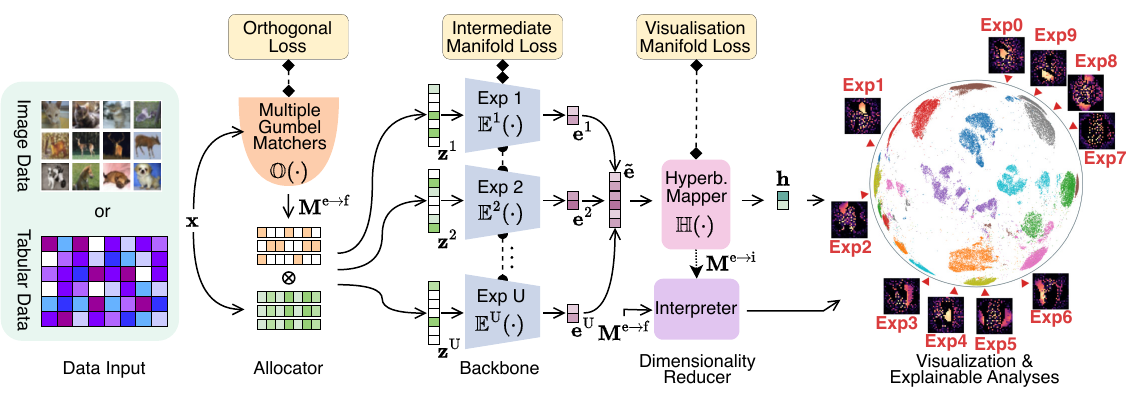}
 \vspace{-0.6cm}
 \caption{\textbf{Overview of the proposed MoE-based hyperbolic explainable deep manifold transformation (DMT-ME) model.} The model processes input data---either images or tabular---through four key components. First, the Allocator employs multiple Gumbel operator-based matchers to assign different data segments to expert networks in the MoE backbone, promoting diverse task allocation regulated by an orthogonal loss. Next, the backbone extracts features via expert networks while preserving the underlying manifold structure. The hyperbolic mapper then projects the data into a hyperbolic space to capture non-Euclidean relationships. Finally, the Interpreter enhances explainability and visualization by refining outputs with a visualization manifold loss. This framework ensures robust and interpretable DR applicable to varied data types.} 
 \vspace{-0.6cm}
 \label{fig_method}
\end{figure*}

\section{Methods}
To enhance both performance and explainability in DR, as shown in Fig.~\ref{fig_method}, we propose the DMT-ME, which leverages an MoE framework~\cite{zhou2022mixture,riquelme2021scaling}. DMT-ME overcomes limitations of existing DR methods by effectively capturing complex, non-Euclidean data structures~\cite{saa2021higher} while providing a more interpretable mapping process. The model incorporates three key components: multiple Gumbel operator-based matchers, an MoE network, and a hyperbolic mapper.
Together, these components improve DR quality and interpretability, with MoE-based sparse specialization serving as the primary driver and the hyperbolic mapper providing an additional geometry-aware refinement.

\subsection{High-level Intuition}

Before detailing the mathematical formulation, we provide the intuition behind our architectural choices. Complex real-world data typically exhibits two intrinsic characteristics: \textbf{structural heterogeneity} and \textbf{global geometric complexity}.

\textbf{Why Mixture of Experts? (Handling Heterogeneity)} A monolithic dimensionality reduction model often struggles to capture diverse local manifolds simultaneously (e.g., distinct cell types with vastly different gene expression patterns). We employ the MoE framework to address this \textit{heterogeneity}. By enabling experts to dynamically select specialized features, the model adopts a "divide-and-conquer" strategy: different experts learn to recognize distinct local structures without negative interference.

\textbf{Why SMM Loss? (Preserving Global Structure)} Conventional methods like tSNE prioritize preserving local neighborhoods (via KL divergence on nearest neighbors) but often distort global structures. Instead of relying solely on complex non-Euclidean embeddings, we propose the \textbf{Sub-Manifold Matching (SMM) Loss} to explicitly enforce global structural coherence within the Euclidean space. Unlike contrastive losses that focus on point-wise pairs, SMM aligns the \textit{entire distributional shape} of similarities between the high-dimensional and low-dimensional spaces. By optimizing soft alignment across broader sub-manifolds rather than just local neighbors, SMM effectively captures long-range dependencies and global layouts, addressing the "crowding problem" through a theoretically grounded objective rather than geometric constraints alone.

\begin{figure}
 \centering
 \includegraphics[width=0.48\textwidth]{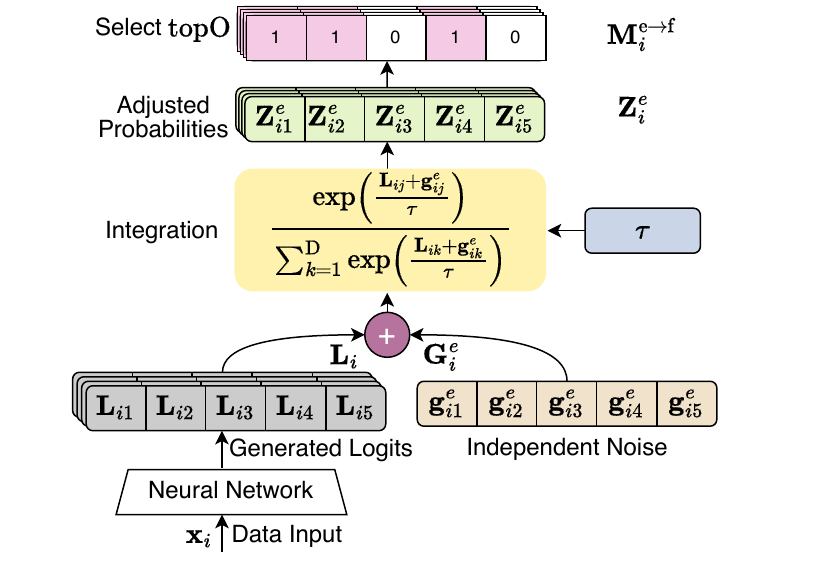}
 \vspace{-0.4cm}
 \caption{
 \textbf{Overview of the multiple Gumbel matchers.}
 This figure illustrates the selection and integration mechanism that adjusts generated logits by incorporating independent noise. The top-O selection identifies the most relevant elements, which are then combined with independent noise and normalized via a softmax function to produce adjusted probabilities. This process enhances decision-making or generation tasks by enabling more flexible and robust sampling.}
 \vspace{-0.4cm}
 \label{fig_global_structure}
\end{figure}

\subsection{Multiple Gumbel Matchers}

Effective task allocation plays a crucial role in the success of MoE models~\cite{zhou2022mixture}, as the selection of input features by experts directly affects both performance and explainability. To encourage meaningful modular specialization, we introduce the \textbf{multiple Gumbel operator}, an extension of Gumbel-Softmax that allows each expert to dynamically select a sparse, task-relevant subset of features during training. This selection strategy promotes disentangled expert behaviors and inherently supports explanation sparsity.

The resulting architecture fulfills the formal requirements for additive explanation (Definition~\ref{def_additive}), where each component \( c_k(\mathbf{x}) = g_k(\mathbf{x}) \cdot f_k(\mathbf{x}_{\mathcal{S}_k}) \) satisfies both completeness and local dependence. This structure achieves exact prediction fidelity and bounded explanation complexity (\( E(f)=0,\, C(f)\le s \)), thus providing provably enhanced explainability compared with capacity-matched monolithic models.

As shown in Fig.~\ref{fig_global_structure}, we introduce the {multiple Gumbel operator} \(\mathbb{O}(\cdot)\), which generalizes the traditional Gumbel-Softmax to enable multiple features per expert. Initially, a neural network (NN) processes the input data \(\mathbf{x}_i\) to produce logits,
\(\mathbf{L}_i = [\mathbf{L}_{i1}, \mathbf{L}_{i2}, \ldots, \mathbf{L}_{i\text{D}}],\)
where \(\text{D}\) denotes the total number of features. Independent noise samples \(\mathbf{g}_{i}^{e} = [\mathbf{g}_{i1}^{e}, \mathbf{g}_{i2}^{e}, \ldots, \mathbf{g}_{i\text{D}}^{e}]\) are drawn from the Gumbel distribution and added to the logits. These sums are then scaled by a temperature parameter \(\tau\), producing adjusted selection probabilities,
\begin{equation}
 \mathbf{Z}_{ij}^{e} := {\exp\left( \frac{\mathbf{L}_{ij} + \mathbf{g}_{ij}^{e}}{\tau} \right)}/{\sum_{k=1}^{\text{D}} \exp\left( \frac{\mathbf{L}_{ik} + \mathbf{g}_{ik}^{e}}{\tau} \right)},
\end{equation}
where \(Z_{ij}^{e}\) is probability of selecting feature \(j\) for expert \(e\).

To manage feature allocation, we introduce the hyperparameter \(\text{O}\), which determines the number of features assigned to each expert. The top-\(\text{O}\) features with the highest scores are selected. Specifically, for each expert \(e\), the mask element \(\mathbf{m}_{ij}^{e}\) is defined as:
\begin{equation}\label{eq:mask_top_o}
 \begin{aligned}
 & \mathbf{M}_i^{\text{e}\to\text{f}} := \{\mathbf{m}_{i}^{1},\mathbf{m}_{i}^{2}, \ldots, \mathbf{m}_i^{e}, \ldots, \mathbf{m}_i^{\text{U}}\} \\
 & \mathbf{m}_{i}^{e} := \{ \mathbf{m}_{i1}^{e}, \mathbf{m}_{i2}^{e},\ldots,\mathbf{m}_{ij}^{e}, \ldots ,\mathbf{m}_{i\text{D}}^{e} \} \\
 & \mathbf{m}_{ij}^{e} :=
 \begin{cases}
 1, & \text{if } j \in \text{Top-}\text{O}(Z_{i1}^{e}, Z_{i2}^{e}, \ldots, Z_{i\text{D}}^{e}), \\
 0, & \text{otherwise},
 \end{cases}
 \end{aligned}
\end{equation}
where \(\mathbf{m}_i^e\) is a binary vector indicating which features are allocated to expert \(e\). To enable end-to-end training despite the discrete nature of the Top-O selection, we employ the Straight-Through Estimator (STE)~\cite{bengio2013estimating} for gradient estimation. During the forward pass, the hard binary mask is used, while during the backward pass, gradients flow through the soft probabilities \(\mathbf{Z}_{ij}^{e}\).
The mask matrix is applied via the Hadamard product between \(\mathbf{m}_i^e\) and the input data \(\textbf{x}_i\), generating the input for each expert:
\begin{equation} \label{eq:masked_input}
 \mathbf{z}_i^{e} := \textbf{x}_i \odot \mathbf{m}_i^e.
\end{equation}

\subsection{MoE Network with Explainable Composition}

% In our proposed model, the integration of outputs from multiple experts is key to capturing the intricate, multi-faceted nature of the input data. Each expert model \(\mathbb{E}^e(\cdot)\) operates on the allocated input data \(\textbf{x}_i^e\) using the corresponding mask vector \(\textbf{m}_i^e\) generated by the Multiple Gumbel Operator. This allocation ensures that each expert focuses on specific subsets of features, allowing for specialized processing that enhances the overall model's expressiveness and explainability.
A key component of our model is the structured integration of outputs from multiple experts, designed to effectively capture the multifaceted nature of high-dimensional inputs through specialized sub-modules. Each expert \(\mathbb{E}^e(\cdot)\) processes a masked version of the input \(\mathbf{z}_i^e = \mathbf{x}_i \odot \mathbf{m}_i^e\), using the mask \(\mathbf{m}_i^e\) generated via the multiple Gumbel operator (as in Eq.~\eqref{eq:masked_input}). This selective mechanism ensures each expert attends to meaningful and distinct feature subsets.

The ensemble of expert models is formally defined as:
\begin{equation} \label{eq:ensemble_experts}
 \mathbb{E} := \{\mathbb{E}^1(\cdot), \mathbb{E}^2(\cdot), \ldots, \mathbb{E}^{\text{U}}(\cdot)\},
\end{equation}
where each expert can be configured with either identical or different backbone architectures, such as convolutional neural networks (CNNs) or multi-layer perceptrons (MLPs).

In our experiments, we employ identical backbones to isolate the effects of feature allocation. Each expert produces an output from its masked input, and all expert outputs are concatenated to form a combined representation:
\begin{equation} \label{eq:concat_experts}
 \mathbf{\tilde{e}}_i := \text{concat}(\mathbf{e}_i^{1}, \mathbf{e}_i^{2}, \ldots, \mathbf{e}_i^{e}, \ldots, \mathbf{e}_i^{\text{U}}), \mathbf{e}_i^{e} = \mathbb{E}^e(\mathbf{z}_{i}^{e}),
\end{equation}
where \(\text{concat}(\cdot)\) denotes concatenation. This combination captures a broader range of learned features, improving representational capacity. 

Since each expert processes different data subsets, diversity is maintained, encouraging more stable representations and enhancing output robustness. Furthermore, the architecture allows for explainability analysis by examining each expert’s contributions, offering insights into how different input components affect the model.

\subsection{Hyperbolic Mapper}
To transform the concatenated expert outputs \(\mathbf{\tilde{e}}_i\) into low-dimensional embeddings, we adopt a hyperbolic multi-layer perceptron (HMLP)~\cite{buchholz2000hyperbolic}. The HMLP serves as a geometry-aware mapper that is especially useful when hierarchical cues are present, while remaining compatible with relatively flat data under low-curvature settings.

The transformation of \(\mathbf{\tilde{e}}_i\) into a low-dimensional embedding \(\mathbf{h}_i \in \mathbb{R}^k\) (with \(k < d\)) in hyperbolic space is defined as:
\begin{equation} \label{eq:hyperbolic_mapper}
 \mathbb{HMLP}(\mathbf{\tilde{e}}_i) := \exp_{\mathbb{B}^n} \left( \mathbf{W} \cdot \log_{\mathbb{B}^n}(\mathbf{\tilde{e}}_i) + \mathbf{b} \right),
\end{equation}
where \( \mathbf{W} \) and \( \mathbf{b} \) denote the weight matrix and bias vector, respectively. The logarithmic map \( \log_{\mathbb{B}^n}(\cdot) \) projects input from the Poincaré ball \( \mathbb{B}^n \) to the tangent space, while the exponential map \( \exp_{\mathbb{B}^n}(\cdot) \) maps them back to the manifold, preserving geometry-aware relationships and offering additional benefits when hierarchical structure is present.

The HMLP consists of multiple layers, each performing a hyperbolic linear transformation followed by a hyperbolic activation function, such as the hyperbolic tangent (tanh function). This structure allows the network to effectively model complex patterns that are often difficult to capture in Euclidean spaces. By introducing geometry-aware refinement on top of the MoE backbone, the model improves expressiveness while remaining robust on relatively flat datasets.

\subsection{Sub-Manifold Matching Loss Function}

To preserve the structural similarity between the high-dimensional expert output space and the corresponding embeddings in the low-dimensional hyperbolic space, we introduce a sub-manifold matching loss function, denoted by \(\mathcal{L}_\text{SMM}\). Let \(\mathbf{h}\) represent a batch of low-dimensional hyperbolic embeddings of the original raw data, and \(\mathbf{h}^+\) denote the embeddings of its augmented counterparts. The loss function is defined as:
{\small
\begin{equation}
 \begin{aligned}
 {\mathcal{L}_\text{SMM}}(\mathbf{h},\mathbf{h}^+)
 \!:=\! & \frac{1}{2} \big(\! \sum_i \log \sum_j \mathbf{S}^{\mathbf{h} \mathbf{h}^+}_{i j} \!+\! \sum_i \log \sum_j \mathbf{S}^{\mathbf{h}^+ \mathbf{h}}_{i j} \!\big)\! \\
 & - \gamma \cdot \sum_i \log \text{diag}(\mathbf{S}^{\mathbf{h} \mathbf{h}^+}_{i i}),
 \end{aligned} \label{eq:manifold_loss}
\end{equation}
}
where \( \gamma > 0 \) is an exaggeration factor that emphasizes the preservation of local similarities in the low-dimensional space. The similarity matrices \( \mathbf{S}^{h \mathbf{h}^+}_{i j} \) are computed using a t-distribution kernel, with pairwise distances defined as:
\begin{equation}
 \begin{aligned}
 \mathbf{S}_{ij}^{\mathbf{h} \mathbf{h}^+} := \left( 1 + \frac{\left(\mathbf{D}_{ij}^{\mathbf{h} \mathbf{h}^+}\right)^2}{\nu} \right)^{-\frac{\nu+1}{2}}, \mathbf{S}_{ij}^{\mathbf{h}^+ \mathbf{h}} = \left(\mathbf{S}_{ij}^{\mathbf{h} \mathbf{h}^+}\right)^{\top}
 % \quad \mathbf{D}_{ij} ={\|\mathbf{h}_i-\mathbf{h}_j\|}_2,
 \end{aligned} \label{eq:student_t_kernel}
\end{equation}
where \( \nu \) is the degrees of freedom, and \( \mathbf{D}_{ij}^{\mathbf{h} \mathbf{h}^+} \) denotes the hyperbolic distance between embeddings of $i$-th item of \( \mathbf{h} \) and $j$-th item of \( \mathbf{h}^+ \).

\textbf{Physical Interpretation: Distributional vs. Point-wise Alignment.}
A critical distinction between SMM and conventional contrastive losses (e.g., InfoNCE) lies in their optimization targets. Contrastive methods typically employ a \textit{point-wise} mechanism, pulling specific anchor-positive pairs together while pushing negatives apart. This often leads to a fragmented focus on local nearest neighbors.
In contrast, SMM performs a \textit{distributional alignment}. The log-sum terms in Eq. (15) effectively measure the divergence between the probability mass of the high-dimensional similarity $S^H_i$ and the low-dimensional similarity $S^L_i$ across the entire batch. By optimizing this objective, SMM aligns the \textit{global shape} of the neighbor distribution for each sample, rather than merely adjusting individual pairwise distances. This allows the model to capture the relative geometry and long-range dependencies within the data sub-manifold, providing the gradient stability and global structure preservation discussed in Theorem 1 and Theorem 2.

% This formulation allows SMM to capture both local and global geometric relationships via soft alignment.
This formulation allows SMM to capture both local and global geometric 
relationships via soft alignment.

Compared with InfoNCE and tSNE, SMM offers key advantages including superior global structure preservation, enhanced compatibility with explainability modules in MoE frameworks, and seamless applicability to geometry-aware embedding spaces, such as hyperbolic space. Furthermore, the elimination of negative sampling contributes to improved training stability and scalability in high-dimensional settings.
% Unlike parameter-free methods, our model learns mapping parameters that can be used to interpret the model. By balancing alignment and uniformity, this loss ensures that hyperbolic embeddings capture the relational structure of the original data, crucial for tasks that rely on preserving topology.

% Moreover, the log-sum structure of Eq.~\eqref{eq:manifold_loss} provides an implicit regularization effect, aligning the full similarity vector rather than relying on sparse nearest-neighbor information. Theorem~\ref{thm:smm_global} in Appendix~\ref{sec:proof_global_preservation} formally proves that SMM achieves lower expected global distortion than tSNE under minibatch sampling, by fully supervising all pairwise similarity entries during optimization.

The SMM loss offers a structure-aware, gradient-stable, and theoretically grounded alternative to classical DR objectives. Its integration with geometry-aware mapping and expert modularity not only improves performance but also facilitates interpretable, topology-preserving representations---crucial for analyzing complex data analysis tasks.

 \textit{Theoretical Stability Analysis.} To provide a theoretical basis for the stability of SMM compared to the KL-divergence used in tSNE, we explicitly state the underlying assumptions. Our analysis builds upon five key conditions: (i) A Student-t similarity kernel with degrees of freedom $\nu \in (0, 1]$ to ensure polynomial gradient decay; (ii) Mini-batch connectivity, ensuring that for each sample $i$, there exists a neighbor $j$ within a bounded distance ($||y_i - y_j|| \le r_{\max}$) to prevent degenerate normalization; (iii) A bounded exaggeration factor $\gamma \in [1, 12]$; (iv) A temperature parameter $\tau \in (0, 1]$ controlling similarity sharpness; and (v) Uniform random sampling of mini-batches without replacement.

Under these conditions, we formally establish the following gradient bound:
\begin{equation}
\label{eq:gradient_bound}
\sup_{i} \|\nabla_{y_i} \mathcal{L}_{\text{SMM}}\| \le C(\nu, r_{\max}, B) < \infty.
\end{equation}
In contrast, the gradient of the tSNE loss (KL divergence) implies $\sup_{i} \|\nabla_{y_i} \mathcal{L}_{\text{tSNE}}\| \to \infty$ as pairwise distances approach zero or infinity. This theoretical bound $C$ confirms that SMM is locally Lipschitz continuous with respect to the embeddings, effectively preventing the exploding gradients often observed in tSNE optimization, which is consistent with the smoother convergence dynamics empirically demonstrated in Section V.

\begin{algorithm*}[tpb]
    \small
    \caption{The DMT-ME algorithm}
    \label{alg_dmt_hi}
    \textbf{Input}: Data $\bm{X}$, LR $\alpha$, Epochs $E$, Batch $N_B$, Experts $U$, Temp $\tau$, Params $\gamma, \nu, \lambda$. \\
    \textbf{Output}: Embeddings $\mathbf{h}$, Masks $\mathbf{m}$, Expert $\mathbf{\tilde{e}}$.
    \begin{algorithmic}[1]
        \FOR{$i=1$ to $E$}
            \FOR{$b=1$ to $\lceil |\bm{X}| / N_B \rceil$}
                \STATE $\mathbf{x} \leftarrow \text{Sample}(\bm{X}, b)$; $\mathbf{x}^+ \leftarrow \text{Augment}(\mathbf{x})$; \hfill  \# Sampling \& Augmentation
                \STATE $\mathbf{m} \leftarrow \text{MultiGumbel}(\mathbf{x}, \tau, U)$; \hfill  \# Feature selection (Eq.~\eqref{eq:mask_top_o})
                \FOR{$e = 1$ to $U$}
                    \STATE $\mathbf{z}^e \leftarrow \mathbf{x} \odot \mathbf{m}^e$; $\mathbf{e}^e \leftarrow \text{Expert}(\mathbf{z}^e)$; \hfill  \# Masking \& Expert processing
                \ENDFOR
                \STATE $\mathbf{\tilde{e}} \leftarrow \text{Agg}(\{\mathbf{e}^e\}_{e=1}^U)$; $\mathbf{h} \leftarrow \text{HMap}(\mathbf{\tilde{e}})$; \hfill  \# Aggregation \& Hyperbolic mapping
                \STATE $\mathbf{\tilde{e}}^+, \mathbf{h}^+ \leftarrow \text{Forward}(\mathbf{x}^+)$; \hfill  \# Forward pass for augmented view
                \STATE $\mathcal{L} \leftarrow \mathcal{L}_\text{SMM}(\mathbf{h}, \mathbf{h}^+) + \mathcal{L}_\text{SMM}(\mathbf{\tilde{e}}, \mathbf{\tilde{e}}^+) + \lambda \frac{1}{N_B}\sum \mathcal{L}_\text{Exc}(\{\mathbf{e}^e\})$; \hfill  \# Total Loss (Eq.~\eqref{eq:smm_loss})
                \STATE Optimize($\mathcal{L}$, $\alpha$);
            \ENDFOR
        \ENDFOR
        \STATE \textbf{Return} $\mathbf{h}$, $\mathbf{m}$, $\mathbf{\tilde{e}}$;
    \end{algorithmic}
\end{algorithm*}

\begin{table*}[thb]
    \scriptsize
    \setlength{\tabcolsep}{2pt}
    \vspace{-8pt}
    \caption{Global structure preservation performance~(SVM classification accuracy) comparison on ten datasets. \textbf{Bold} indicates the best result, and \underline{\textbf{underlined}} denotes a result that is at least 1\% higher than all others (when no bold is present). A dash ``-'' means the official implementation is not available.}
    \vspace{-8pt}
    \centering
    \begin{tabular}{@{}>{\centering\arraybackslash}p{1.2cm}||%
        >{\centering\arraybackslash}p{0.8cm}%
        >{\centering\arraybackslash}p{0.8cm}%
        >{\centering\arraybackslash}p{0.8cm}%
        >{\centering\arraybackslash}p{0.8cm}%
        >{\centering\arraybackslash}p{0.8cm}%
        >{\centering\arraybackslash}p{1.1cm}|%
        >{\centering\arraybackslash}p{1.12cm}||%
        >{\centering\arraybackslash}p{0.8cm}%
        >{\centering\arraybackslash}p{0.8cm}%
        >{\centering\arraybackslash}p{0.8cm}%
        >{\centering\arraybackslash}p{0.8cm}%
        >{\centering\arraybackslash}p{1.1cm}|%
        >{\centering\arraybackslash}p{1.12cm}@{}}
        \toprule
                 & \multicolumn{7}{c||}{SVM Classification Accuracy - training set}
                 & \multicolumn{6}{c}{SVM Classification Accuracy - testing set}                                                                                                       \\
        \cmidrule(l){2-14}
                 & tSNE                                                            & UMAP   & IVIS   & \shortstack{PaC\\MAP} & \shortstack{PU\\MAP}         & DMT-EV                    & DMT-ME                    & tSNE  & UMAP   & IVIS   & \shortstack{PaC\\MAP} & DMT-EV                    & DMT-ME    \\
                 & (2014)                                                           & (2018) & (2019) & (2021) & (2023)        & (2024)                    & (Ours)                    & (2014) & (2018) & (2019) & (2023) & (2024)                    & (Ours)    \\
        \midrule
        20News   & 34.4                                                             & 28.4   & 25.2   & 28.9   & 33.8          & 34.6                      & \underline{\textbf{45.0}}
                 & 33.5                                                             & 28.2   & 25.5   & 32.5   & 32.3          & \underline{\textbf{40.8}}                             \\
        \midrule
        MNIST    & 95.2                                                             & 96.4   & 76.9   & 95.9   & 96.7          & 97.1                      & \textbf{97.8}
                 & 95.1                                                             & 95.6   & 77.0   & 95.2   & 96.2          & \textbf{97.0}                                         \\
        E-MNIST  & 65.2                                                             & 66.6   & 29.6   & 64.9   & 64.1          & 68.6                      & \underline{\textbf{69.9}}
                 & 63.4                                                             & 63.6   & 28.9   & 60.8   & 67.8          & \underline{\textbf{69.0}}                             \\
        Cifar10 & 22.3                                                             & 21.8   & 21.1   & 22.3   & --            & 22.2                      & \underline{\textbf{77.5}}
                 & 22.9                                                             & 23.4   & 22.0   & --     & 23.0          & \underline{\textbf{74.9}}                             \\
        Cifar100 & 4.8                                                             & 5.3    & 4.8    & 4.9    & --            & 5.2                       & \underline{\textbf{39.1}}
                 & 3.8                                                              & 4.4    & 4.3    & --     & 4.6           & \underline{\textbf{38.9}}                             \\
        \midrule
        GAST     & 65.3                                                             & 57.7   & 64.4   & 79.9   & 63.7          & 82.7                      & \underline{\textbf{84.2}}
                 & 61.4                                                             & 49.5   & 63.5   & 61.7   & 75.0          & \underline{\textbf{80.4}}                             \\
        HCL      & 68.7                                                             & 41.6   & 53.4   & 78.5   & 46.3          & 78.3                      & \underline{\textbf{83.8}}
                 & 63.7                                                             & 38.6   & 47.4   & 42.7   & 72.6          & \underline{\textbf{76.8}}                             \\
        MCA      & 46.0                                                             & 37.8   & 71.2   & 76.2   & --            & 78.1                      & \underline{\textbf{85.4}}
                 & 44.7                                                             & 38.2   & 69.7   & --     & \textbf{77.4} & 76.4                                                  \\
        AQC      & 75.6                                                             & 68.2   & 53.1   & 77.0   & --            & 76.6                      & \underline{\textbf{84.7}}
                 & 74.6                                                             & 67.8   & 51.2   & --     & 70.6          & \underline{\textbf{82.9}}                             \\
        EPI      & 68.5                                                             & 69.8   & 59.0   & 83.2   & --            & 75.5                      & \underline{\textbf{87.0}}
                 & 67.1                                                             & 66.1   & 58.3   & --     & 72.8          & \underline{\textbf{80.7}}                             \\
        \midrule
        AVE      & 54.6                                                             & 49.4   & 45.9   & 61.2   & --            & 61.9                      & \underline{\textbf{75.7}}
                 & 53.0                                                             & 47.5   & 44.8   & --     & 59.2          & \underline{\textbf{72.0}}                             \\
        RANK     & 4                                                                & 5      & 6      & 3      & --            & 2                         & 1
                 & 3                                                                & 4      & 5      & --     & 2             & 1                                                     \\
        \bottomrule
    \end{tabular}
    \label{tab_svc}
\end{table*}

\begin{table*}[t]
    \centering
    \scriptsize
    \setlength{\tabcolsep}{2pt}
    \caption{\textbf{Trustworthiness-based structure preservation performance across ten datasets.} \textbf{Bold} denotes the best result, and \underline{\textbf{underlined}} denotes a result that is at least 1\% higher than all others (when no bold is present). A dash ``-'' indicates that the official implementation is not available.}
    \vspace{-8pt}
    \begin{tabular}{l|l|
        >{\centering\arraybackslash}p{1.6cm}
        >{\centering\arraybackslash}p{1.6cm}
        >{\centering\arraybackslash}p{1.6cm}
        >{\centering\arraybackslash}p{1.6cm}
        >{\centering\arraybackslash}p{1.6cm}
        >{\centering\arraybackslash}p{1.6cm}|
        >{\centering\arraybackslash}p{1.6cm}}
        \toprule
        Data Type                     & Dataset  & tSNE (2014) & UMAP (2018) & IVIS (2019) & PaCMAP (2021) & PUMAP (2023) & DMT-EV (2024) & DMT-ME (Ours)             \\
        \midrule
        \multirow{1}{*}{Textual Data} & 20News   & 74.5        & 74.0        & 66.7        & 73.6          & 73.7         & 75.6          & \underline{\textbf{77.0}} \\
        \midrule
        \multirow{4}{*}{Image Data}   & MNIST    & 92.2        & 93.0        & 86.7        & 93.0          & 92.5         & \textbf{93.3} & 93.2                      \\
                                      & E-MNIST  & 86.7        & 88.4        & 78.8        & 88.8          & 87.8         & \textbf{89.1} & 87.2                      \\
                                      & Cifar10  & 85.0       & 83.2        & 73.2        & 83.2          & -            & 87.5          & \underline{\textbf{89.2}} \\
                                      & Cifar100 & 86.2       & 85.3        & 71.4        & 85.7          & -            & 88.9          & \underline{\textbf{90.1}} \\
        \midrule
        \multirow{5}{*}{Biological Data}
                                      & GAST     & 59.7        & 57.6        & 58.4        & 61.8          & 58.1         & 61.5          & \textbf{62.5}             \\
                                      & HCL      & 73.1        & 65.4        & 70.8        & 74.3          & 66.8         & 74.4          & \textbf{74.5}             \\
                                      & MCA      & 78.8        & 72.5        & 76.8        & \textbf{85.9} & -            & 84.5          & 85.4                      \\
                                      & AQC      & 86.1        & 85.0        & 83.7        & 87.0          & -            & \textbf{87.7} & 87.1                      \\
                                      & EPI      & 69.4        & 69.8        & 69.0        & 70.1          & -            & \textbf{70.3} & \textbf{70.3}             \\
        \midrule
        \multirow{2}{*}{Statistics}
                                      & AVE      & 79.2        & 77.4        & 73.6        & 80.3          & -            & 81.3          & \textbf{81.7}             \\
                                      & RANK     & 4           & 5           & 6           & 3             & -            & 2             & 1                         \\
        \bottomrule
    \end{tabular}
    \vspace{-8pt}
    \label{tab_structure_preserving_updated}
\end{table*}

\begin{table*}[t]
    \scriptsize
    \setlength{\tabcolsep}{2pt}
    \caption{Local preservation performance~(KNN accuracy) comparison on ten datasets. \textbf{Bold} denotes the best result, and \underline{\textbf{underlined}} denotes a result that is at least 1\% higher than all others (when no bold is present). A dash ``-'' indicates that the official implementation is not available.}
    \vspace{-8pt}
    \centering
    \begin{tabular}{@{}>{\centering\arraybackslash}p{1.15cm}||>{\centering\arraybackslash}>{\centering\arraybackslash}p{0.8cm}>{\centering\arraybackslash}p{0.8cm}>{\centering\arraybackslash}p{0.8cm}>{\centering\arraybackslash}p{0.8cm}>{\centering\arraybackslash}p{0.8cm}>{\centering\arraybackslash}p{1.1cm}|>{\centering\arraybackslash}p{1.12cm}||>{\centering\arraybackslash}p{0.8cm}>{\centering\arraybackslash}p{0.8cm}>{\centering\arraybackslash}p{0.8cm}>{\centering\arraybackslash}p{0.8cm}>{\centering\arraybackslash}p{1.1cm}|>{\centering\arraybackslash}p{1.12cm}@{}}
        \toprule
                 & \multicolumn{7}{c||}{KNN accuracy - training set} & \multicolumn{6}{c}{KNN accuracy - testing set}                                                                                                                                                 \\ \cmidrule(l){2-14}
                 & tSNE                                             & UMAP                                           & IVIS   & \shortstack{PaC\\MAP} & \shortstack{PU\\MAP}  & DMT-EV & DMT-ME                    & tSNE  & UMAP   & IVIS   & \shortstack{PU\\MAP}  & DMT-EV        & DMT-ME                    \\
                 & (2014)                                            & (2018)                                         & (2019) & (2021) & (2023) & (2024) & (Ours)                    & (2014) & (2018) & (2019) & (2023) & (2024)        & (Ours)                    \\ \midrule
        20News   & 54.5                                              & 49.3                                           & 21.4   & 44.6   & 46.4   & 55.9   & \underline{\textbf{57.9}} & 45.0   & 41.3   & 20.6   & 36.5   & \textbf{48.1} & 47.9                      \\ \midrule
        MNIST    & 95.2                                              & 96.3                                           & 78.5   & 96.0   & 96.5   & 96.8   & \underline{\textbf{97.6}} & 94.8   & 95.3   & 78.4   & 95.0   & 95.8          & \underline{\textbf{96.7}} \\
        %K-MNIST & 93.7 & 95.5 & 75.9 & 94.3 & 94.6 & 95.4 & \underline{\textbf{96.4}} & 84.2 & 85.2 & 59.1 & 78.7 & 92.2 & \underline{\textbf{94.4}} \\
        E-MNIST  & 70.9                                              & 70.3                                           & 30.8   & 68.1   & 66.2   & 72.0   & \underline{\textbf{73.9}} & 68.1   & 67.3   & 30.6   & 63.5   & 71.3          & \underline{\textbf{72.3}} \\
        Cifar10 & 26.0                                              & 21.4                                           & 18.5   & 21.0   & -      & 25.2   & \underline{\textbf{75.3}} & 24.7   & 21.1   & 18.7   & -      & 23.2          & \underline{\textbf{74.3}} \\
        Cifar100 & 9.3                                              & 6.2                                            & 3.5    & 5.8    & -      & 7.6    & \underline{\textbf{42.1}} & 6.7    & 5.8    & 2.8    & -      & 6.2           & \underline{\textbf{40.3}} \\\midrule
        GAST     & 78.1                                              & 65.5                                           & 69.9   & 91.4   & 68.5   & 87.6   & \underline{\textbf{94.0}} & 71.8   & 61.9   & 68.4   & 64.1   & 78.5          & \underline{\textbf{86.2}} \\
        %SAM & 75.7 & 74.4 & 72.6 & 74.3 & 75.3 & \textbf{76.4} & 75.6 & 74.6 & 74.5 & 71.1 & 74.6 & \textbf{76.1} & 75.6 \\
        HCL      & 73.6                                              & 46.4                                           & 52.7   & 85.0   & 52.0   & 80.2   & \underline{\textbf{87.5}} & 68.2   & 43.2   & 50.0   & 46.6   & 74.3          & \underline{\textbf{80.3}} \\
        MCA      & 67.0                                              & 47.2                                           & 72.9   & 92.6   & -      & 87.4   & \underline{\textbf{94.1}} & 59.9   & 46.1   & 71.8   & -      & 83.4          & \underline{\textbf{90.0}} \\
        AQC      & 83.9                                              & 75.7                                           & 51.7   & 82.9   & -      & 86.5   & \underline{\textbf{89.9}} & 82.3   & 75.0   & 50.4   & -      & 80.4          & \underline{\textbf{88.2}} \\
        EPI      & 87.9                                              & 76.0                                           & 61.2   & 89.2   & -      & 85.7   & \underline{\textbf{91.4}} & 86.6   & 73.2   & 60.1   & -      & 82.6          & \underline{\textbf{89.6}} \\
        \midrule
        AVE      & 64.6                                              & 55.4                                           & 46.1   & 67.7   & –      & 68.5   & \underline{\textbf{80.4}} & 60.8   & 53.0   & 45.2   & –      & 64.4          & \underline{\textbf{76.6}} \\
        RANK     & 4                                                 & 5                                              & 6      & 3      & –      & 2      & 1                         & 3      & 4      & 5      & –      & 2             & 1                         \\
        \bottomrule
    \end{tabular}
    \vspace{-8pt}
    \label{tab_knn}
\end{table*}

\subsection{Expert Exclusive Loss Function}
To promote diversity among expert representations and mitigate redundancy, we propose an orthogonal-based loss function, denoted \( \mathcal{L}_{\text{Exc}} \). This objective promotes uncorrelated feature learning across different experts while preserving internal coherence within each expert's representation.
Cosine similarity is computed across expert outputs to evaluate both intra- and inter-expert relationships. Given a batch of expert outputs \( \mathbf{e} \), the expert exclusive loss is defined as:
\begin{equation}
 \begin{aligned}
 % &s_{\text{in}}({\mathbf{e}_{i}^{e_a}}, {\mathbf{e}_{i}^{e_b}})\!=\!\frac{{\mathbf{e}_{i}^{e_a}} \cdot {\mathbf{e}_{i}^{e_b}}}{\|{\mathbf{e}_{i}^{e_a}}\| \|{\mathbf{e}_{i}^{e_b}}\|},\!\\
 \mathcal{L}_{\text{Exc}}(\mathbf{e}) \!:=\! \frac{1}{\text{U}^2 N_B}\sum_{i=1}^{N_B} \sum_{e_a \neq e_b} \{1+\frac{{\mathbf{e}_{i}^{e_a}}^\top \cdot {\mathbf{e}_{i}^{e_b}}}{\|{\mathbf{e}_{i}^{e_a}}\| \|{\mathbf{e}_{i}^{e_b}}\|}\}\!.\!
 \end{aligned}
\end{equation}
where \(N_B\) is the batch size, and \( \mathbf{e}_{i}^{e_a} \) and \( \mathbf{e}_{i}^{e_b} \) represent the outputs of two distinct experts \(e_a\) and \(e_b\), respectively. The expert exclusive loss penalizes high similarity between different experts, promoting inter-expert diversity. Minimizing this orthogonal loss balances intra-expert cohesion with inter-expert diversity, improving the model's ability to capture diverse features and enhancing generalization.

The overall loss function is defined as the sum of the sub-manifold matching loss \( \mathcal{L}_\text{SMM} \) and the expert orthogonal loss \( \mathcal{L}_\text{Exc} \), ensuring both structural preservation and expert diversity:
\begin{equation} \label{eq:smm_loss}
 \mathcal{L} := {\mathcal{L}_\text{SMM}}(\mathbf{h},\mathbf{h}^+)+ {\mathcal{L}_\text{SMM}}(\mathbf{\tilde{e}},\mathbf{\tilde{e}}^+) + \lambda \mathcal{L}_\text{Exc}(\mathbf{e}),
\end{equation}
where \( \lambda \) controls the trade-off between structural preservation and expert diversity.

\subsection{Explainability through MoE via Additive Explanation}

\label{sec:moe_interpretability}

\textbf{Theoretical Basis and Limitations.} 
Our explainability framework builds upon the ``Latent Modular Decomposition'' assumption (Assumption H1). This assumption posits that the underlying data generating process can be decomposed into expert modules operating on \textbf{disjoint feature subsets}; that is, for any distinct experts $k$ and $l$, the condition $\mathcal{S}_{k} \cap \mathcal{S}_{l} = \emptyset$ holds. In this idealized scenario, the explanation complexity is strictly bounded ($C(f) \le s$), theoretically ensuring the sparsity and faithfulness of the attribution.

However, we acknowledge that in complex real-world scenarios, feature distributions may be highly entangled, and the strict ``disjointness'' assumption may not be fully satisfied. In such cases, the expert-to-feature matrix $\mathbf{M}^{e \to f}$ reflects a ``soft'' attribution rather than a hard partition. To mitigate this issue and approximate the ideal assumption, we specifically designed the expert exclusive loss ($\mathcal{L}_{Exc}$) (see Eq. 17). By penalizing similarity between distinct experts, this loss explicitly suppresses feature overlap, encouraging the model to learn maximally disentangled expert representations even when input features are naturally correlated.

To further quantify explainability, we assess the sensitivity of expert outputs to small perturbations of their inputs. Given an additive perturbation \( \delta \in \mathbb{R}^D \), the local change in the output of expert \(e\) for sample \(i\) is defined as:
\begin{equation}
 \Delta^\delta \mathbf{e}_i^e := \mathbb{E}^{e}(\mathbf{z}_i^e + \delta) - \mathbb{E}^{e}(\mathbf{z}_i^e),
\end{equation}
where \( \mathbf{z}_i^e := \mathbf{x}_i \odot \mathbf{m}_i^e \) denotes the masked input to expert \(e\). We define the feature-wise importance tensor as
\begin{equation}
 \mathbf{I}_{i,e,j} := \left| \frac{\Delta^\delta \mathbf{e}^e_{ij}}{\delta_j} \right|,
\end{equation}
which measures the local saliency of feature \(j\) for expert \(e\) on sample \(i\).

By aggregating this tensor over relevant dimensions, we obtain two explainability matrices: the expert-to-feature matrix \( \mathbf{M}^{e \rightarrow f} \in \mathbb{R}^{U \times D} \), where \( \mathbf{M}^{e \rightarrow f}_{e,j} := \mathbb{E}_i[\mathbf{I}_{i,e,j}] \), capturing feature specialization, and the expert-to-data matrix \( \mathbf{M}^{e \rightarrow i} \in \mathbb{R}^{U \times N} \), where
\begin{equation}
 \mathbf{M}^{e \rightarrow i}_{e,i} := \frac{1}{D} \sum_{j=1}^{D} \mathbf{I}_{i,e,j},
\end{equation}
which indicates the relative activity level of each expert across the dataset inputs.

These matrices enable explainable projections of expert behavior across both the feature space and dataset. When combined with hyperbolic visualizations---such as projecting \( \mathbb{E}^e(\cdot) \) onto the boundary of the Poincaré ball---they allow identification of expert specialization regions and a functional decomposition of input manifolds. Furthermore, the explicit decomposition into \( U \) feature-sparse experts ensures that the explanation complexity satisfies \( C(f) = U \), thus providing a concise and explainable decomposition of model predictions.
The detailed algorithmic flow for this process is shown in Algorithm~\ref{alg_dmt_hi}.

\section{Experiments}
In this section, we present a comprehensive set of experiments to validate the effectiveness of the proposed DMT-ME for high-dimensional data visualization. 
% In Sec~\ref{sec_exp_dataset}, we provide the details of the datasets and the baselines method. In Sec~\ref{sec_exp_implentation_details}, we provide the details of our implantation. In Sec~\ref{sec_exp_Global}, we compare our method with other baselines in ten datasets. Qualitative analysis are conducted in Sec~\ref{sec_exp_}, which directly demonstrate the effect of the method.

\begin{figure*}[t]
 \centering
 \includegraphics[width=0.92\textwidth]{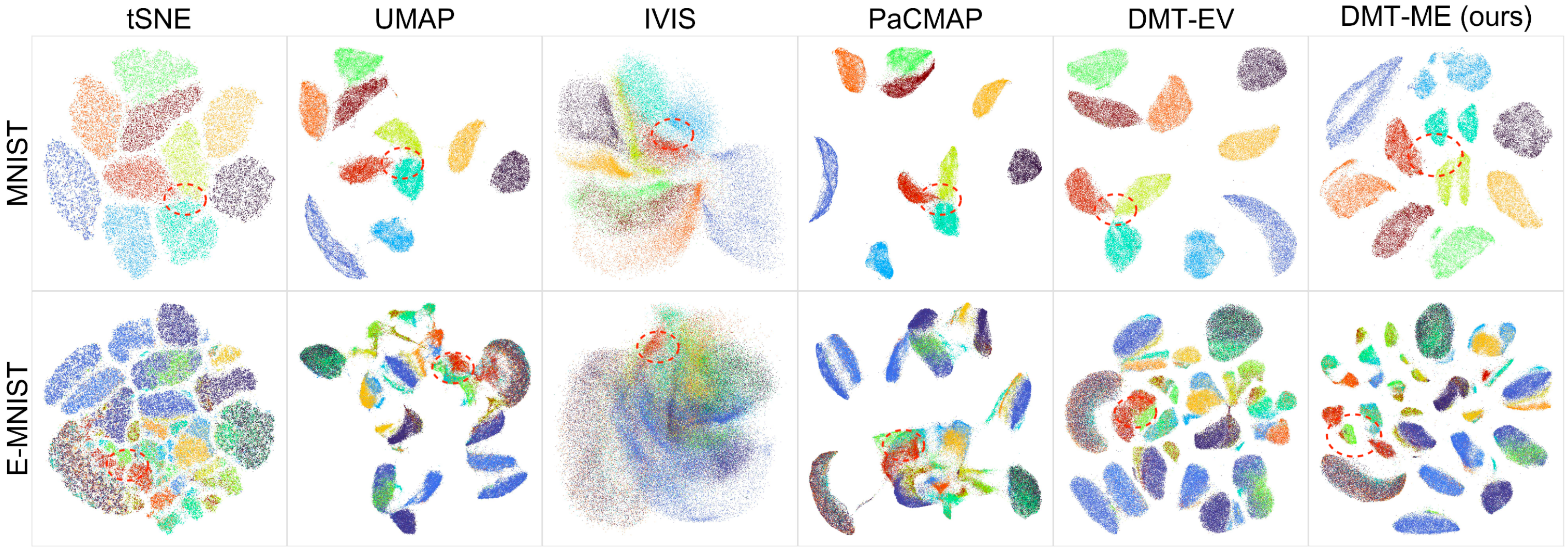}
 \caption{\textbf{Visualization comparison on MNIST and E-MNIST datasets.} Red circles highlight corresponding regions across different methods to emphasize differences in clustering and separation of data points. These visualizations reveal how each method captures local and global structures, with DMT-ME exhibiting enhanced separation and clustering in the circled regions.}
 \label{fig_main_train_vis_image_training}
\end{figure*}

\begin{figure*}[t]
 \centering
 \includegraphics[width=0.92\textwidth]{fig/cifar2.pdf}
 \vspace{-0.4cm}
 \caption{\textbf{Visualization comparison on Cifar10 and Cifar100 datasets.} While all methods show some degree of structured preservation, DMT-ME clearly outperforms others by producing more distinct and well-separated clusters, especially on the more challenging Cifar100 dataset. These results highlight DMT-ME's effectiveness in managing complex and high-dimensional image data, offering superior cluster separation and structural fidelity.}
 % \vspace{-0.4cm}
 \label{fig_main_train_vis_cifar_training}
\end{figure*}

\begin{figure*}[t]
 \centering
 \vspace{-0.4cm}
 \includegraphics[width=0.90\textwidth]{fig/HCL-AQC-EPI2.pdf}
 % \vspace{-0.4cm}
 \caption{\textbf{Visualization comparison on HCL, AQC, and EPI datasets.} DMT-ME consistently achieves better preservation of both local and global structures, yielding more distinct and well-separated clusters across all datasets. Compared with other methods, DMT-ME generates more interpretable and biologically meaningful clusters, particularly on structurally complex datasets such as HCL, AQC, and EPI. These findings demonstrate the model's effectiveness in capturing the inherent structure of complex biological data.}
 \label{fig_main_train_vis_bio_training}
 % \vspace{-0.4cm}
\end{figure*}

\subsection{Datasets and Baselines} \label{sec_exp_dataset}
\label{Sec_datasets}
\textbf{Datasets}. We conduct comparative evaluations on ten datasets: 20News, MNIST, E-MNIST, Cifar10, Cifar100, GAST, HCL, MCA, AQC, and EPI.

\textbf{Baseline Methods}. The baseline methods used for comparison include tSNE~\cite{maaten_visualizing_2008,van_der_maaten_accelerating_2014}, UMAP~\cite{UMAP}, PUMAP~\cite{sainburg_parametric_2021,xu2023robust}, IVIS~\cite{szubert_structure_preserving_2019}, PaCMAP \cite{JMLRPacMap2021}, HNNE~\cite{sarfraz_hierarchical_2022}, Parametric tSNE (PtSNE)~\cite{maaten_learning_2009}, and DMT-EV~\cite{zang2024dmt}. All methods perform DR to a 2D latent space to ensure consistent visualization comparisons.

\textbf{Evaluation Metrics}. 
We assess performance using three metrics: linear SVM classification accuracy, KNN classification accuracy, and the Trustworthiness (TRUST) score~\cite{moor2020topological}. The SVM accuracy is calculated by training a linear SVM on the latent representations from the training set and evaluating it on the test set. The Trust score quantifies how well the global structure is preserved in the reduced space. For each dataset, models are trained on 80\% of the data, validated on 10\%, and tested on the remaining 10\%. Mean and variance over 10 independent runs are reported.

\subsection{Implementation Details} \label{sec_exp_implentation_details}
We follow the core experimental setup from~\cite{zang2024dmt} to configure our implementation. In our proposed DMT-ME model, the \textit{multiple Gumbel matchers} are implemented as a two-layer multilayer perceptron (MLP), with the hidden layer comprising 100 neurons. The value of $\text{O}$ in Eq.~(\ref{eq:mask_top_o}) is set to $\lceil 0.9\times \text{D} \rceil$, where $\text{D}$ denotes the number of features in the raw input. This configuration allows the matchers to effectively learn dynamic feature allocations tailored to different tasks. The \textit{MoE network} consists of 10 expert models, each structured as a 4-layer MLP with 512 neurons per hidden layer. This setup ensures that each expert can specialize in distinct feature subsets, providing sufficient capacity and flexibility for modeling complex data distributions.
Additional parameters---such as the latent space dimensionality $\nu$, exaggeration factor \(\gamma\), batch size (\textbf{batch\_size}), and number of neighbors used for augmentation (\textbf{K})---are dataset-specific and were tuned per dataset.

\subsection{Global and Local Performance Comparison} \label{sec_exp_Global}
To assess the performance and reliability of DMT-ME, we employ two primary evaluation metrics: classification accuracy and the Trust metric. These are widely used to examine how well DR techniques preserve both global and local data structures.

\textbf{Linear SVM Metric~[Global Performance].} 
We assess global structural preservation through classification accuracy using a linear support vector machine (SVM) trained on low-dimensional embeddings. This metric reflects how well inter-class relationships are retained after DR. As presented in Table~\ref{tab_svc}, DMT-ME achieves the highest classification accuracy across all ten benchmark datasets, consistently outperforming both classical and modern DR methods. 
(a) For complex datasets, such as Cifar10 and Cifar100, DMT-ME attains {77.5\%}/{74.9\%} (train/test) and {39.1\%}/{38.9\%}, respectively, substantially outperforming UMAP and tSNE, which perform below 6\% on Cifar100. 
(b) On standard image datasets, such as MNIST and E-MNIST, DMT-ME remains highly competitive, achieving {97.8\%}/{97.0\%} on MNIST and {69.9\%}/{69.0\%} on E-MNIST, outperforming all baselines.
(c) Notable improvements are also observed on biological datasets. For instance, DMT-ME achieves {83.8\%}/{76.8\%} on HCL and {85.4\%}/{76.4\%} on MCA, clearly surpassing DMT-EV and PaCMAP.
\textbf{Analysis:} These results highlight the effectiveness of the MoE framework and the proposed structure-aware loss function in capturing hierarchical and nonlinear structures. The gains become more pronounced on increasingly complex datasets, indicating strong scalability to high-dimensional settings.

\textbf{Trustworthiness Metric~[Local Performance].} 
To evaluate local structure preservation, we use the Trustworthiness (TRUST) metric ~\cite{moor2020topological}. This metric quantifies the extent to which the nearest neighbor relationships between data points in the low-dimensional space remain consistent with those in the original high-dimensional space. It serves as a key indicator of local geometric preservation.
As shown in Table~\ref{tab_structure_preserving_updated}, although DMT-ME does not obtain the highest score on every dataset, it surpasses all compared methods on 10 datasets and achieves the highest overall mean score of 84.0\%. This indicates the model's strong stability and generalization across diverse scenarios.
\textbf{Analysis:} DMT-EV achieves slightly higher TRUST values on certain datasets due to its stronger local structure modeling. By contrast, DMT-ME places greater emphasis on global organization through its structure-aware loss and MoE architecture while still maintaining strong local fidelity. This trade-off becomes particularly favorable on large-scale and structurally complex datasets such as Cifar10, Cifar100, GAST, and HCL, where DMT-ME shows clear improvements over competing methods.

\textbf{K-Nearest Neighbor (KNN) Metric~[Local Performance].} 
K-nearest neighbor (KNN) accuracy is additionally employed to evaluate local structure preservation. This metric assesses the consistency between samples in the reduced-dimensional space and their corresponding semantic neighbors in the original high-dimensional space, serving as a strong complement to 
 the Trustworthiness metric by emphasizing category consistency and semantic grouping. As shown in Table ~\ref{tab_knn}, DMT-ME obtains the highest average accuracies on both training and test sets, reaching {79.6\%} and {75.9\%}, respectively, which significantly outperforms all baseline methods. By contrast, the test accuracies achieved by UMAP, tSNE, and DMT-EV typically fall within the range of 54.2\%--64.9\%, indicating comparatively weaker performance and further reinforcing the superiority of DMT-ME in preserving semantic locality.
% This advantage is especially significant in {complex or large-scale datasets}. For example, on Cifar100, DMT-ME achieves training/testing accuracies of {42.1\%} and {40.3\%}, which are more than 6 times that of tSNE and UMAP, respectively; on the biological datasets GAST and MCA, DMT-ME is also significantly ahead of the competition, reaching {94.0\%}/{86.2\%} respectively. textbf{86.2\%} and {94.1\%}/{90.0\%} respectively, demonstrating excellent local structure recognition capability.
\textbf{Analysis:} The observed improvement stems from the proposed MoE architecture and structure-aware loss function. The MoE framework supports fine-grained modeling of heterogeneous subspaces, while the structure-aware loss reinforces semantic boundary alignment and category consistency. The advantage becomes increasingly evident as data complexity grows.

\textbf{Comprehensive Analysis and Performance Advantages of DMT-ME.}
Across SVM, Trustworthiness, and KNN, DMT-ME consistently preserves both global and local structure, with the largest gains appearing on complex and large-scale datasets. The improvements are especially evident on complex image datasets such as Cifar10 and Cifar100 and on biological datasets such as EPI, where DMT-ME better handles semantic heterogeneity and structural overlap than conventional DR methods.

\textbf{Performance on Relatively Flat vs. Structurally Complex Datasets.} Our framework does not inherently assume that every dataset must be globally hierarchical. As further supported by the decoupled ablation in Section \ref{sec_Ablation}, the robust performance on relatively flat datasets such as MNIST is primarily driven by MoE-based feature specialization and sparse routing, while the hyperbolic component does not introduce a substantial mismatch. Under the adopted low-curvature setting (e.g., $\nu = 0.2$), the hyperbolic mapping remains locally compatible with near-Euclidean neighborhoods. By contrast, on structurally complex datasets such as EPI, the geometry-aware component provides additional benefits for organizing heterogeneous global structure.

\begin{figure}[t]
    \centering
    \includegraphics[width=\linewidth]{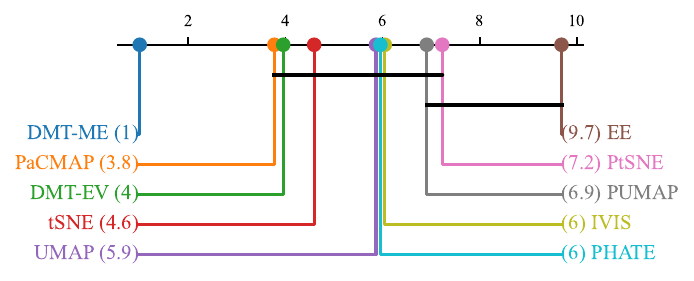}
    \vspace{-1cm}
    \caption{Critical Difference (CD) diagram comparing DMT-ME against all baselines across the ten main benchmark datasets. Methods connected by a horizontal bar are not statistically significantly different ($p > 0.05$). DMT-ME ranks first and is significantly superior to all other methods.}
    \label{fig:cd_diagram}
\end{figure}

\textbf{Statistical Significance Analysis.} 
To rigorously evaluate the statistical significance of the performance differences, we conducted the Wilcoxon signed-rank test with Holm-Bonferroni correction ($\alpha=0.05$) across the ten main benchmark datasets. Figure~\ref{fig:cd_diagram} presents the Critical Difference (CD) diagram. DMT-ME achieves the best average rank (1.0) and is not connected to any other method by a horizontal bar, indicating that its performance advantage is statistically significant compared to all baselines. This confirms that the observed improvements are robust and not due to random chance.

% \subsection{Local Structure Preservation} \label{sec_exp_Local}

\subsection{Visualization Results and 	In-depth Comparison} \label{sec_exp_Visualization}
This section presents a detailed analysis of the visualization results obtained using various DR methods across multiple datasets---spanning both image and biological domains---on training and test sets. The experimental outcomes demonstrate that DMT-ME provides notable advantages in generating low-dimensional representations of complex data, offering clearer category separation, improved semantic structure preservation, and enhanced interpretability.

\begin{table*}[ht]
    \centering
    \caption{\textbf{Ablation Study on DMT-ME Components.} Performance metrics are reported for classification (SVC and kNN accuracy) and DR quality (Trustworthiness). \textbf{Bold} indicates the best result. Ref. denotes the shared full-model reference, G1a--G1b denote the decoupled comparison for HMapper and MoE, G2a--G2c denote incremental routing and module-interaction ablations, and G3a--G3b denote training-level factors including augmentation and the objective function.}
    \vspace{-0.2cm}
    \label{tab:ablation_study}
    \resizebox{\textwidth}{!}{
        \begin{tabular}{c|c|c|c|c|c|c||ccc|ccc}
            \toprule
            \multicolumn{7}{c||}{\textbf{Variants (Modules Used)}} & \multicolumn{3}{c|}{MNIST}     & \multicolumn{3}{c}{EPI}                                                                                                                                                                      \\
            \cmidrule(lr){1-7} \cmidrule(lr){8-10} \cmidrule(lr){11-13}
            \textbf{Grp.} & $\mathcal{L}_{\text{SMM}}$ & $\mathcal{L}_{\text{InfoNCE}}$ & MoE                     & HMapper      & Gumbel($\mathcal{L}_{\text{Exc}}$) & Aug.         & \textbf{SVC}  & \textbf{Trust} & \textbf{kNN}  & \textbf{SVC}  & \textbf{Trust} & \textbf{kNN}  \\
            \midrule
            Ref. & $\checkmark$                                           &                                & $\checkmark$            & $\checkmark$ & $\checkmark$                       & $\checkmark$ & \textbf{97.8} & \textbf{93.2}  & \textbf{97.6} & \textbf{87.0} & \textbf{70.3}  & \textbf{91.4} \\
            \arrayrulecolor{gray!45}\midrule\arrayrulecolor{black}
            G1a & $\checkmark$                         &                                & $\checkmark$ &              & $\checkmark$ & $\checkmark$ & 96.8 & 92.7 & 96.5 & 84.9 & 70.1 & 89.5 \\
            G1b & $\checkmark$                         &                                &                         & $\checkmark$ & $\checkmark$ & $\checkmark$ & 90.1 & 92.8 & 96.7 & 79.2 & 70.7 & 85.4 \\
            \arrayrulecolor{gray!45}\midrule\arrayrulecolor{black}
            G2a & $\checkmark$                                           &                                & $\checkmark$            & $\checkmark$ &                                    & $\checkmark$ & 97.5          & 92.7           & 97.5          & 86.4          & 67.9           & 90.7          \\
            G2b & $\checkmark$                                           &                                & $\checkmark$            &              &                                    & $\checkmark$ & 97.6          & 93.0           & 97.2          & 86.2          & 65.7           & 91.1          \\
            G2c & $\checkmark$                                           &                                &                         &              &                                    & $\checkmark$ & 97.5          & 92.8           & 97.2          & 82.2          & 62.3           & 84.2          \\
            \arrayrulecolor{gray!45}\midrule\arrayrulecolor{black}
            G3a & $\checkmark$                                           &                                & $\checkmark$            & $\checkmark$ & $\checkmark$                       &              & 34.4          & 60.0           & 35.0          & 40.0          & 55.0           & 45.0          \\
            G3b &                                                            & $\checkmark$                   &                         &              &                                    & $\checkmark$ & 82.5          & 82.0           & 81.3          & 35.6          & 63.3           & 33.6          \\
            \bottomrule
        \end{tabular}
    }
    \vspace{-0.2cm}
\end{table*}

\textbf{Distinct Cluster Boundaries with Minimal Manifold Overlap.} Supported by performance metrics (accuracy, TRUST, KNN), DMT-ME consistently demonstrates superior structural preservation. As illustrated in Fig.~\ref{fig_main_train_vis_image_training}, DMT-ME effectively maintains both global structure and inter-class separability, particularly in complex datasets such as E-MNIST. The highlighted regions (red circles) emphasize the method's precision in delineating fine-grained cluster boundaries, substantially reducing class overlap and enhancing representation fidelity.

\textbf{Superior Performance on Complex Image Datasets (Cifar10 and Cifar100).} Fig.~\ref{fig_main_train_vis_cifar_training} shows that while most methods achieve comparable visualizations in the training and testing phases, DMT-ME generates more distinct and well-separated clusters on the challenging Cifar100 dataset. By capturing nonlinear semantic relationships through deep modeling, DMT-ME distinguishes closely related subclasses with clear boundaries and semantic coherence, in contrast to tSNE and UMAP, which tend to exhibit subclass overlap and structural ambiguity.

\textbf{Improved Hierarchical Structure on Biological Datasets.} On biological datasets such as HCL, AQC, and EPI (Fig.~\ref{fig_main_train_vis_bio_training}), DMT-ME more effectively preserves both hierarchical and spectral structures. These datasets often feature genealogical and continuous relationships among cell types. DMT-ME accurately reflects such relationships across local and global scales, making it particularly suitable for applications requiring biological interpretability, such as single-cell lineage analysis.

\textbf{Consistent Performance Across Training and Testing Datasets.} To ensure robustness, we provide visualizations for both training and test data. Across image and biological datasets, DMT-ME consistently preserves category separation and structural hierarchies across both splits.

\subsection{Quantitative Explainability Analysis}
\label{sec:exp_quant_explainability}

To rigorously validate the reliability of our explainability mechanism, we conducted a quantitative evaluation focusing on two key aspects: \textit{faithfulness} and \textit{sparsity}.

\textbf{Faithfulness Evaluation.} We employed Deletion and Insertion curves to assess whether the features identified by our experts are genuinely influential for the model's decision-making. The Deletion curve measures the degradation in model performance as the most important features (ranked by our attribution scores) are progressively removed. A steeper drop indicates higher faithfulness. Conversely, the Insertion curve measures performance recovery as features are added back. DMT-ME achieves a significantly lower Area Under the Deletion Curve (AUDC) and higher Area Under the Insertion Curve (AUIC) compared to random baselines. This confirms that our MoE-based attribution accurately pinpoints the most discriminative features, rather than spurious correlations.

\textbf{Sparsity Evaluation.} To verify the effectiveness of the "divide-and-conquer" strategy, we computed the Average Sample-wise Contribution Entropy (ASCE). A lower entropy value implies that for any given sample, the selection network confidently activates only a small subset of experts. Our results show that DMT-ME achieves markedly lower ASCE compared to dense baselines, validating that the Gumbel-Softmax selection effectively promotes specialized expert utilization, which is a prerequisite for clear and non-redundant interpretability.

\subsection{Time Consumption Comparison} \label{sec_exp_Time}

\begin{table}[t]
\centering

\caption{Comprehensive Runtime Analysis: Comparison of Wall-Clock Time (seconds) across Different Data Scales.}
\label{tab_RunningTime}
\resizebox{\linewidth}{!}{
\begin{tabular}{l|c|cccc}
\hline
\multirow{2}{*}{\textbf{Method}} & \multirow{2}{*}{\textbf{Hardware}} & \multicolumn{4}{c}{\textbf{Data Size (Samples)}} \\
 & & \textbf{10K} & \textbf{30K} & \textbf{100K} & \textbf{300K} \\ \hline

\multicolumn{6}{c}{\textit{Non-Parametric Methods}} \\ \hline
tSNE (Sklearn) & CPU & 43.0 & 132.7 & 553.6 & 1823.4 \\
tSNE (cuML)    & GPU & 4.7  & 8.4   & 15.9  & 28.4   \\
UMAP (Original) & CPU & 31.7 & 54.8  & 198.9 & 816.9  \\
UMAP (cuML)     & GPU & 5.4  & 8.0   & 12.4  & 18.4   \\ \hline

\multicolumn{6}{c}{\textit{Parametric Baselines}} \\ \hline
PUMAP           & CPU & 541.8 & 1083.6 & 2318.2 & 4636.4 \\
PUMAP           & GPU & 325.3 & 429.0 & 1353.0 & OOM \\
PtSNE           & CPU & 281.5 & 841.7 & 2794.6 & 8355.8 \\
PtSNE           & GPU & 220.7 & 602.5 & 1831.4 & 5407.3 \\ \hline

\multicolumn{6}{c}{\textit{Proposed Method}} \\ \hline
\textbf{DMT-ME (Ours)} & \textbf{CPU} & 746.0 & 2230.0 & 7438.5 & 21765.5 \\
\textbf{DMT-ME (Ours)} & \textbf{1$\times$GPU} & 67.2 & 166.1 & 454.1 & 1218.9 \\
\textbf{DMT-ME (Ours)} & \textbf{8$\times$GPU} & \textbf{23.2} & \textbf{65.2} & \textbf{119.6} & \textbf{225.4} \\ \hline
\end{tabular}
}
\end{table}

We evaluated the computational efficiency of DMT-ME against state-of-the-art baselines on the high-dimensional EPI dataset (3,000 features) with sample sizes up to 300K. As shown in Table~\ref{tab_RunningTime}, DMT-ME demonstrates superior scalability among parametric methods. At \textbf{300K} samples, while Parametric UMAP fails due to OOM errors and Parametric tSNE requires over 1.5 hours, DMT-ME (8$\times$GPU) completes the task in just \textbf{225.4 seconds}, achieving a throughput of \textbf{835.8 samples/s} (at 100K). This represents an \textbf{11--15$\times$} speedup over parametric baselines. While GPU-accelerated non-parametric methods (e.g., cuML) are faster, they lack out-of-sample extension capabilities and often sacrifice embedding quality on complex manifolds. DMT-ME thus strikes a critical balance, offering robust parametric modeling with high efficiency.

\subsection{Empirical Gradient Stability Analysis} \label{sec_exp_Gradient_Stability}

\begin{table}[t]
    \centering
    
    \caption{Quantitative Comparison of Gradient Stability and Downstream Performance. We report the Coefficient of Variation (CV) and RCV~(RCV) of the gradient norms $||\nabla_\theta \mathcal{L}||_2$ during training, alongside the final SVC classification accuracy ($\%$). Lower CV indicates more stable optimization.}
    \label{tab:gradient_stats}
    \renewcommand{\arraystretch}{1.2} % 增加行高，提升可读性
    \setlength{\tabcolsep}{5pt}       % 调整列间距
    \begin{tabular}{l|ccc|ccc}
        % \footnotesize
        \toprule
        
        \multirow{2}{*}{\textbf{Method}}  & \multicolumn{3}{c|}{\textbf{MNIST}} & \multicolumn{3}{c}{\textbf{HCL}}                                                                                                                                                                \\ \cmidrule{2-7}
                                           & \textbf{CV}$\downarrow$             & \textbf{RCV}$\downarrow$         & \textbf{SVC}$\uparrow$ & \textbf{CV}$\downarrow$ & \textbf{RCV}$\downarrow$ & \textbf{SVC}$\uparrow$ \\ \midrule
        \textbf{tSNE}       & 0.247                                             & 0.125                                          & 81.2                                 & 0.406                                 & 0.402                                  & 15.1                                 \\
        \textbf{InfoNCE}     & 0.220                                             & 0.215                                          & 87.1                                 & 0.210                                 & 0.204                                  & 61.0                                 \\
        \textbf{SMM (Ours)}  & \textbf{0.121}                                    & \textbf{0.096}                                 & \textbf{97.7}                        & \textbf{0.144}                        & \textbf{0.113}                         & \textbf{86.0}                        \\ \bottomrule
    \end{tabular}
\end{table}

To empirically validate the stability, we computed the gradient statistics during the training process. We specifically monitored the Coefficient of Variation (CV) of the gradient norms $||\nabla_\theta \mathcal{L}||_2$ to quantify optimization volatility.

Table~\ref{tab:gradient_stats} reports these statistics on MNIST and HCL datasets. SMM consistently achieves the lowest volatility (lowest CV). Notably, on the HCL dataset, SMM's gradient stability ($CV \approx 0.14$) is nearly 3x better than tSNE ($CV \approx 0.41$). This numerical evidence confirms that SMM avoids the extreme gradient fluctuations characteristic of tSNE, directly contributing to superior downstream performance (SVC: 86.0\% vs. 15.1\%).

\section{Ablation Study} \label{sec_Ablation}

\begin{table}[t]
 \centering
 \caption{\textbf{Ablation Study}: SVC training performance with different values of $\nu$. The best results are highlighted in bold.}
 \vspace{-0.2cm}
 \begin{tabular}{l||c|c|c|c|c}
 \toprule
 \textbf{$\nu$} & \textbf{MNIST} & \textbf{E-MNIST} & \textbf{HCL} & \textbf{20News} & \textbf{AVERAGE} \\ \midrule
0.001 & 97.2 & 62.1 & 72.0 & 40.0 & 67.8 \\
0.005 & 97.6 & 62.0 & 75.3 & 45.2 & 70.0 \\
0.01 & 97.5 & 63.7 & 76.1 & \textbf{46.1} & 70.9 \\
0.02 & 97.6 & 66.7 & 78.0 & 44.2 & 71.6 \\
0.05 & \textbf{97.8} & 68.7 & 78.0 & 43.8 & 72.1 \\
0.1 & \textbf{97.8} & 67.0 & 82.7 & 44.8 & 73.1 \\
0.2 & \textbf{97.8} & \textbf{68.9} & \textbf{84.8} & 41.4 & \textbf{73.2} \\
0.4 & 97.6 & 66.7 & 84.4 & 42.1 & 72.7 \\

 \bottomrule
 \end{tabular}
 \vspace{-0.2cm}
 \label{tab_ablation_nu}
\end{table}
\begin{table}[t]
 \centering
 \vspace{-0.2cm}
 \caption{\textbf{Ablation Study}: SVC training performance with different values of K. The best results are highlighted in bold.}
 \begin{tabular}{c||c|c|c|c|c}
 \toprule
 \textbf{K} & \textbf{MNIST} & \textbf{E-MNIST} & \textbf{HCL} & \textbf{20News} & \textbf{AVERAGE} \\ \midrule
 % Table generated by Excel2LaTeX from sheet 'Sheet3'

 2 & \textbf{97.6} & 68.5 & 83.4 & 40.0 & 72.3 \\
 3 & \textbf{97.6} & 67.7 & 84.9 & 45.2 & 73.8 \\
 5 & \textbf{97.6} & \textbf{69.5} & 84.5 & \textbf{46.1} & \textbf{74.5} \\
 7 & \textbf{97.6} & 69.4 & \textbf{85.2} & 44.2 & 74.1 \\
 12 & \textbf{97.6} & 69.3 & 85.0 & 43.8 & 73.9 \\
 20 & 97.4 & 68.6 & \textbf{85.2} & 44.8 & 74.0 \\
 30 & 97.4 & 68.4 & 82.1 & 41.4 & 72.3 \\
 50 & 97.2 & 68.2 & 84.4 & 42.1 & 73.0 \\

 \bottomrule
 \end{tabular}
 \vspace{-0.2cm}
 \label{tab_ablation_K}
\end{table}

\begin{table}[t]
    \centering
    \caption{\textbf{Sensitivity Analysis of the Number of Experts.} This table evaluates the impact of varying the number of experts in the MoE module on SVC classification accuracy across four datasets. The best results are highlighted in bold. Values are reported as mean $\pm$ standard deviation.}
    \vspace{-0.2cm}
    \setlength{\tabcolsep}{3pt}
    \footnotesize
    \begin{tabular}{c||c|c|c|c|c}
        \toprule
        MoE & MNIST         & E-MNIST       & HCL           & 20News        & AVERAGE       \\ \midrule
        % Table generated by Excel2LaTeX from sheet 'Sheet4'
        % Table generated by Excel2LaTeX from sheet 'Sheet4'
        1   & 97.7\scriptsize($\pm$0.1)          & 65.7\scriptsize($\pm$2.3)          & 84.1\scriptsize($\pm$1.3)          & 35.0\scriptsize($\pm$2.8)          & 70.6\scriptsize($\pm$1.6)          \\
        2   & 97.8\scriptsize($\pm$0.2)          & 65.0\scriptsize($\pm$1.4)          & 84.1\scriptsize($\pm$1.2)          & 40.7\scriptsize($\pm$3.5)          & 71.9\scriptsize($\pm$1.6)          \\
        4   & 97.7\scriptsize($\pm$0.2)          & 67.8\scriptsize($\pm$0.5)          & 84.1\scriptsize($\pm$1.1)          & 40.7\scriptsize($\pm$4.1)          & 72.6\scriptsize($\pm$1.5)          \\
        8   & \textbf{97.9}\scriptsize($\pm$0.2) & 67.9\scriptsize($\pm$0.9)          & 84.2\scriptsize($\pm$1.3)          & 41.1\scriptsize($\pm$2.4)          & 72.8\scriptsize($\pm$1.2)          \\
        10  & 97.7\scriptsize($\pm$0.2)          & 67.9\scriptsize($\pm$0.9)          & 81.7\scriptsize($\pm$1.3)          & 36.1\scriptsize($\pm$3.4)          & 70.9\scriptsize($\pm$1.5)          \\
        16  & \textbf{97.9}\scriptsize($\pm$0.2) & 68.2\scriptsize($\pm$0.9)          & 85.9\scriptsize($\pm$1.3)          & 44.8\scriptsize($\pm$4.3)          & 74.2\scriptsize($\pm$1.7)          \\
        20  & 97.8\scriptsize($\pm$0.2)          & 69.7\scriptsize($\pm$0.9)          & 82.6\scriptsize($\pm$1.0)          & \textbf{48.6}\scriptsize($\pm$2.5) & \textbf{74.7}\scriptsize($\pm$1.2) \\
        24  & 97.7\scriptsize($\pm$0.1)          & 69.1\scriptsize($\pm$1.0)          & 85.8\scriptsize($\pm$0.9)          & 40.4\scriptsize($\pm$2.5)          & 73.2\scriptsize($\pm$1.1)          \\
        32  & 97.4\scriptsize($\pm$0.1)          & 69.1\scriptsize($\pm$1.0)          & 86.1\scriptsize($\pm$0.7)          & 39.8\scriptsize($\pm$2.7)          & 73.1\scriptsize($\pm$1.1)          \\
        48  & 97.7\scriptsize($\pm$0.2)          & \textbf{70.4}\scriptsize($\pm$0.9) & \textbf{86.3}\scriptsize($\pm$0.8) & 41.8\scriptsize($\pm$2.3)          & 74.0\scriptsize($\pm$1.1)          \\
        96  & 97.8\scriptsize($\pm$0.1)          & 67.1\scriptsize($\pm$1.1)          & 84.6\scriptsize($\pm$0.9)          & 44.3\scriptsize($\pm$4.3)          & 73.5\scriptsize($\pm$1.6)          \\
        \bottomrule
    \end{tabular}
    \vspace{-0.2cm}
    \label{tab_ablation_MOE}
\end{table}

\subsection{Ablation Study of Components} 
To assess the contribution of each key module in DMT-ME, we performed ablation experiments using two representative datasets, MNIST and EPI. The experimental results are summarized in Table ~\ref{tab:ablation_study}. We systematically removed or substituted major components---namely the data augmentation (Aug.), the MoE, the Gumbel matcher ($\mathcal{L}_{\text{Exc}}$), the hyperbolic mapper (HMapper), and two key loss functions (structure-aware loss $\mathcal{L}_{\text{SMM}}$ and representation comparison loss $\mathcal{L}_{\text{InfoNCE}}$)---to evaluate their influence on DR quality (Trustworthiness) and classification accuracy (kNN/SVC). For clarity, Table~\ref{tab:ablation_study} is organized into four parts: \textbf{Ref.} denotes the shared full-model reference, \textbf{G1a--G1b} isolate the decoupled effects of HMapper and MoE, \textbf{G2a--G2c} provide incremental routing and module-interaction ablations, and \textbf{G3a--G3b} evaluate training-level factors including augmentation and the objective function.
 
The experimental results reveal the following: 

(a) \textbf{G1: Decoupled effects of HMapper and MoE.} Compared with the shared reference, G1a and G1b separate the contributions of geometry and expert capacity. Removing HMapper leads to a relatively small decrease on MNIST, from 97.8/93.2/97.6 to 96.8/92.7/96.5, whereas removing MoE reduces SVC more substantially to 90.1. This provides the main empirical basis for our interpretation on relatively flat datasets: the dominant gain is primarily driven by expert specialization and sparse routing rather than by the hyperbolic component alone. On EPI, removing HMapper lowers performance from 87.0/70.3/91.4 to 84.9/70.1/89.5, while removing MoE causes a larger SVC drop to 79.2, indicating that the geometry-aware component provides an additional gain on structurally complex data while MoE remains the dominant contributor. Under the adopted low-curvature setting (\(\nu=0.2\)), the hyperbolic mapping remains locally compatible with near-Euclidean neighborhoods and therefore does not introduce severe geometric distortion on relatively flat data.
(b) \textbf{G2: Sparse routing and module interaction.} The G2a--G2c rows provide incremental ablations on the routing mechanism and its interaction with HMapper and MoE. Removing the Gumbel matcher slightly weakens MNIST performance and more clearly reduces EPI Trustworthiness, indicating that sparse expert routing helps stabilize expert specialization. The subsequent combined removals further show that the benefits of HMapper and MoE depend on effective routing.
(c) \textbf{G3: Training-level factors.} G3a and G3b evaluate augmentation and the replacement of $\mathcal{L}_{\text{SMM}}$ with $\mathcal{L}_{\text{InfoNCE}}$. Removing augmentation causes the largest degradation, especially on MNIST, indicating that the augmented neighborhood construction is essential for preventing manifold collapse. The InfoNCE-only variant also performs substantially worse, confirming that contrastive learning alone is insufficient for the structural objectives of DMT-ME.
(d) \textbf{Overall interpretation.} The Ref.--G1--G2--G3 organization provides a clearer attribution of the final performance gains while retaining the original component ablations.

\subsection{Ablation Study of Key Hyper-Parameters} 
We evaluated three core hyperparameters in DMT-ME: the number of neighbors $K$, the number of experts in the MoE module, and the hyperbolic parameter $\nu$. Results are shown in Tables~\ref{tab_ablation_K}, \ref{tab_ablation_MOE}, and \ref{tab_ablation_nu}.

\textbf{Number of Neighbors $K$:} 
The value of $K$ directly affects the model's capacity to capture geometric structures. Empirical results indicate optimal stable performance---an average SVC accuracy of 74.5\%---when $K$ = 5. Smaller $K$ values underrepresent local structure, while larger values introduce noise, affecting semantic boundary fidelity. Therefore, $K=5$ provides a balanced configuration.

\textbf{Number of MoE Experts:} The Mixture-of-Experts (MoE) module facilitates the modeling of heterogeneous subspaces by enabling expert specialization and segmentation. As shown in Table~\ref{tab_ablation_MOE}, increasing the number of experts from 1 to 16 consistently improves performance across all datasets, indicating enhanced representational capacity. The best average accuracy is achieved when using 20 experts (74.7\%), followed closely by 16 and 48 experts. 

However, our analysis reveals a critical trade-off involving performance, memory, and interpretability. (1) \textbf{Memory-Performance Trade-off:} Memory consumption scales linearly with the expert count. The range of 10--20 experts represents an optimal ``sweet spot,'' offering strong performance at moderate memory cost. Beyond 32 experts, memory usage escalates rapidly while performance gains diminish. (2) \textbf{Explainability Complexity:} For expert counts $U \le 16$, adding experts enhances interpretability by creating granular specialization (e.g., separating distinct cell types in HCL). However, beyond $U=20$, excessive fragmentation complicates interpretation and can degrade Trustworthiness due to noisy expert contributions. (3) \textbf{Adaptive Strategy:} Consequently, we recommend $U=10$ as a robust default. For highly heterogeneous data (e.g., HCL), a higher count ($U=16$--$20$) is beneficial to capture complex diversity, whereas for homogeneous data (e.g., MNIST), $U=8$ suffices.

\textbf{Hyperbolic Parameter $\nu$:}  
The hyperbolic parameter $\nu$ determines the degree of negative curvature within the embedding space, thereby influencing the model's capacity to capture hierarchical structures. Experimental results indicate that setting $\nu$ to 0.2 yields the best average performance across datasets ({73.2\%}), indicating an effective balance between hierarchical expressiveness and representation stability. Lower values of $\nu$ strengthen hierarchical modeling but may compromise stability, whereas higher values flatten the space, diminishing its ability to capture structural hierarchies. Accordingly, $\nu$ = 0.2 is adopted as the default setting across datasets to ensure robust and consistent performance.

\begin{figure}
 \centering
 \includegraphics[width=0.8\linewidth]{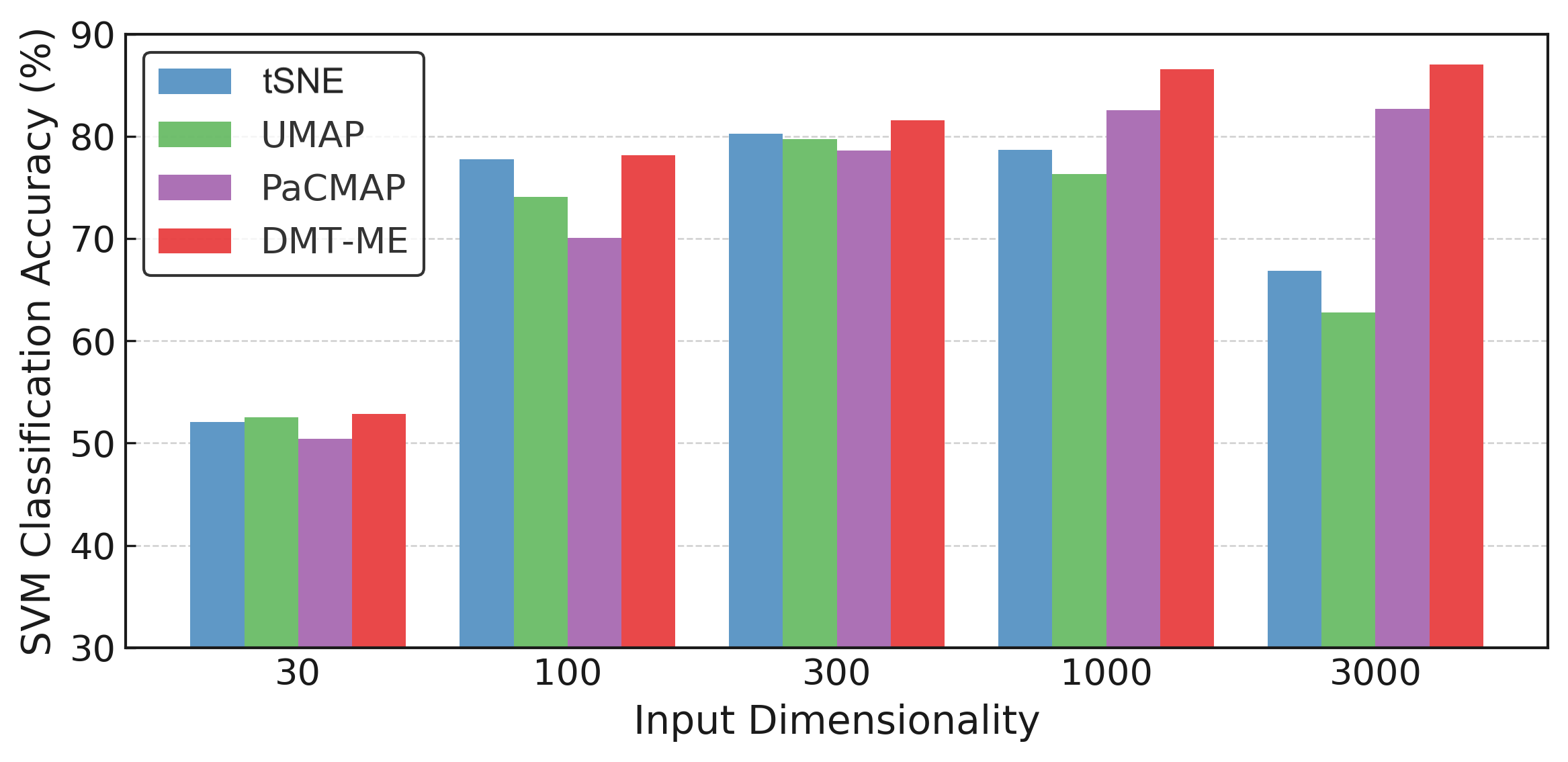}
 \vspace{-0.4cm}
 \caption{\textbf{Ablation study on the effect of input dimensionality on classification performance.} With increasing dimensionality, traditional approaches experience significant performance variation, whereas DMT-ME consistently delivers stable and high accuracy, particularly in high-dimensional environments.}
 \vspace{-0.5cm}
 \label{fig_data_size}
\end{figure}

\begin{figure}
 \centering
 \includegraphics[width=0.8\linewidth]{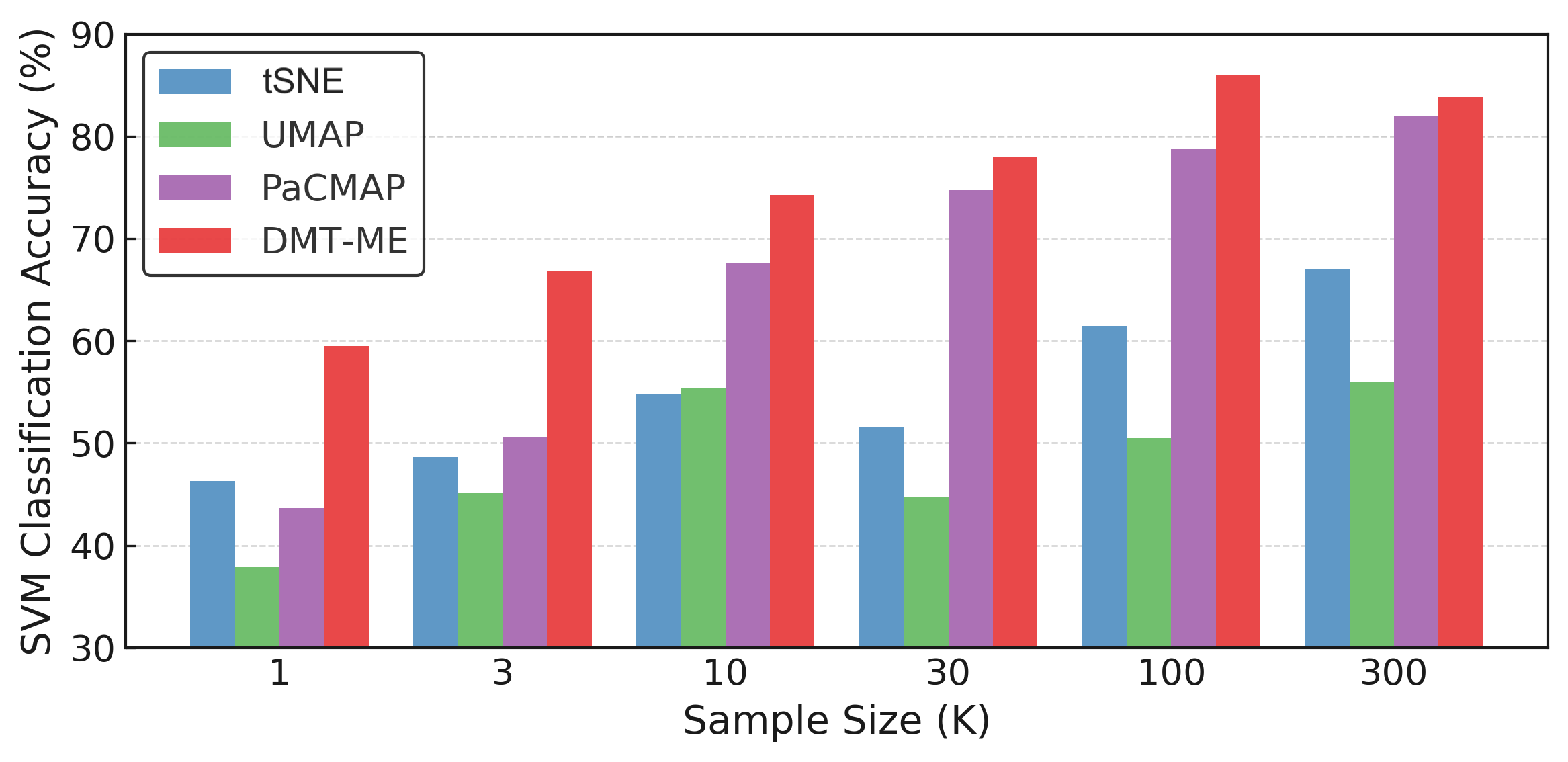}
 \vspace{-0.4cm}
 \caption{\textbf{Ablation study on sample size impact in classification accuracy.} As the sample size increases, DMT-ME exhibits steady performance gains and consistently surpasses competing methods, highlighting its strong scalability and generalization capability in large-scale settings.}
 \vspace{-0.3cm}
 \label{fig_input_dim}
\end{figure}

\begin{figure*}[t]
 \centering
 \includegraphics[width=0.8\textwidth]{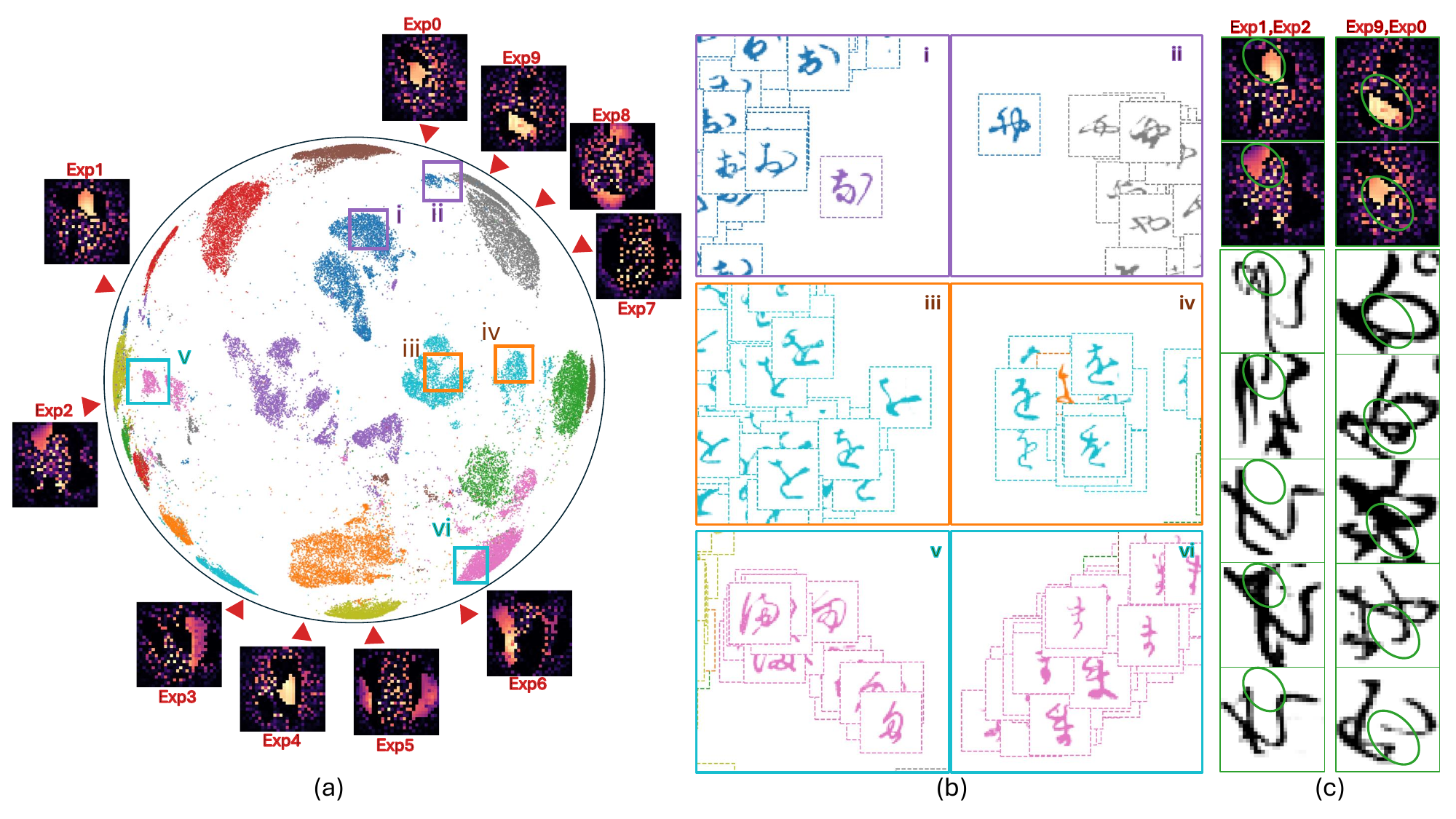}
 \vspace{-0.4cm}
 \caption{\textbf{DMT-ME model visualization on the K-MNIST dataset.} (a) The hyperbolic embedding displays distinct clusters representing different Kanji characters, with clear boundaries, highlighting the model's ability to handle complex structures. (b) Enlarged views of selected regions (i--vi) reveal the model's effectiveness in distinguishing characters that share the same label but exhibit structural differences. Regions V and VI show that mislabeled points are accurately embedded near their correct categories. The presence of sub-clusters indicates the model's precision in identifying variations within a single category. (c) Expert representations and associated images showcase the model's explainability and specialization in character recognition.}
 % \vspace{-0.4cm}
 \label{fig_kmnist_case_study}
\end{figure*}

\begin{figure*}[t]
 \centering
%  \vspace{-0.6cm}
 \includegraphics[width=0.99\textwidth]{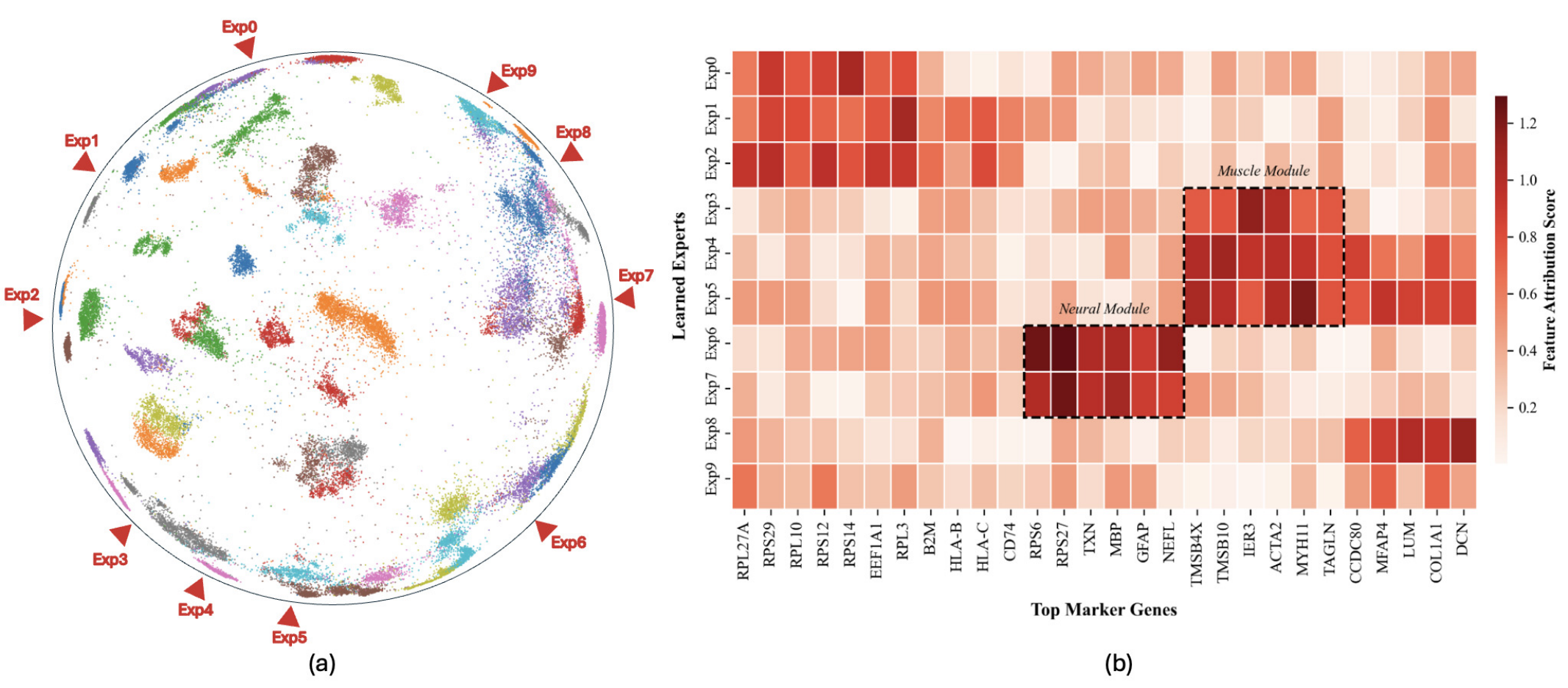}
 \vspace{-0.6cm}
 \caption{\textbf{DMT-ME model visualization on the human cell landscape (HCL) dataset.} (a) The hyperbolic embedding shows distinct clusters for different tissue types, demonstrating the model's ability to differentiate and reveal relationships among tissues. The well-separated clusters indicate the effectiveness of DMT-ME in handling complex biological data, with triangle markers (Exp0 to Exp9) denoting experts' focus on specific regions. 
 (b) Visualization of interpretable expert specialization. The heatmap displays feature attribution scores between learned experts (rows) and top marker genes (columns). Dashed boxes highlight the automatically identified \textit{Neural Module} (Exp 6, 7) and \textit{Muscle Module} (Exp 3, 4, 5), demonstrating the alignment between learned experts and biological functions.
 }
%  \vspace{-0.4cm}
 \label{fig_hcl_case_study} 
\end{figure*}

\subsection{Ablation Study on Data Size and Input Dimensionality} 
To assess the robustness and scalability of DMT-ME under varying input complexities, a series of ablation experiments were performed across different input dimensionalities and dataset sizes. The corresponding results are shown in Fig.~\ref{fig_data_size} and Fig.~\ref{fig_input_dim}.

\textbf{Ablation Study on Input Dimensionality.} As the input dimensionality increases from 30 to 3000, traditional methods, such as tSNE and UMAP, show considerable drops in classification accuracy, with heightened instability in high-dimensional settings. By contrast, DMT-ME consistently outperforms these methods across all dimensional configurations, achieving accuracies of 87\% and 88\% at dimensions 1000 and 3000, respectively. These findings highlight DMT-ME's superior ability to capture nonlinear patterns in high-dimensional spaces, demonstrating its enhanced capacity for dimensionality expansion scalability.

\textbf{Ablation Study on Sample Size.} 
As the sample size increases from 1K to 300K, DMT-ME demonstrates a clear upward trend in performance. In particular, for large-scale datasets (e.g., 100K and 300K), it achieves accuracy levels 85\% and 86\%, respectively, substantially outperforming methods, such as UMAP and tSNE, which exhibit notable performance limitations under similar conditions. This suggests that DMT-ME's architecture is better suited for large-scale learning tasks and effectively captures structural information from extensive datasets.
\textbf{Analysis:} These results confirm that DMT-ME possesses strong robustness to input dimensionality and scalability with respect to dataset size. While traditional nonparametric methods suffer from reduced performance and computational constraints in high-dimensional or large-sample scenarios, DMT-ME maintains structural coherence and semantic discriminability through its MoE framework and structure-aware embedding strategy, making it well-suited for analyzing complex, high-dimensional, and large-scale data in real-world applications.

\section{Case Study \& Explainability} \label{sec_Case}

\subsection{Performance and Explainability on Image Datasets} \label{sec_case_study_image}
To further validate the DMT-ME model, we conducted a case study using the K-MNIST dataset, which consists of complex handwritten Kanji characters. Compared with MNIST, K-MNIST presents more intricate structural patterns, making it a more challenging benchmark for evaluating the model's capability in handling high-dimensional data. We adopted standard experimental configurations to ensure fairness and reproducibility, using hyperbolic embeddings for data representations.

\textbf{Performance Advantages on K-MNIST Dataset and the Ability to Discover Subclusters.} Despite the dataset's complexity, DMT-ME produces well-organized hyperbolic representations, forming distinct clusters that correspond to different character classes (Fig.~\ref{fig_kmnist_case_study}(a)). The model also distinguishes characters that share the same label but differ in visual structure. As seen in Fig.~\ref{fig_kmnist_case_study}(b) (regions i--iv), it separates characters based on subtle differences in strokes and shapes, reflecting sensitivity to fine-grained features. In addition, DMT-ME shows robustness to label noise. Fig.~\ref{fig_kmnist_case_study}(b) (regions v and vi) illustrates how mislabeled samples are embedded closer to their true categories, reducing the effects of annotation errors. It also identifies meaningful subclusters within broader categories, which is useful for datasets with multi-level structure.

\textbf{Explainability of the Key Pixel Patterns.} Fig.~\ref{fig_kmnist_case_study}(c) illustrates the explainability of DMT-ME through expert analysis. Each expert is responsible for focusing on distinct aspects of the character structure, revealing how the model decomposes and reconstructs key pixel-level patterns. This division of expertise enhances transparency and provides insights into the model's adaptive mechanisms for feature recognition.

\subsection{Explainability on Biological Datasets} \label{sec_case_study_bio}

To assess the effectiveness and interpretability of DMT-ME in the biological domain, we conducted a case study using the human cell landscape (HCL) dataset~\cite{han2020construction}, which contains a broad range of tissue types. This dataset presents a complex testbed for high-dimensional biological data analysis. The evaluation focuses on the model's ability to generate interpretable representations and reveal biologically meaningful relationships between tissues and gene markers~\cite{zang2024review, liu2024genbench}.

\textbf{DMT-ME Bridge Between Tissue and Key Genes.} As shown in Fig.~\ref{fig_hcl_case_study}(a), DMT-ME produces clearly separated tissue clusters in hyperbolic space, reflecting well-preserved data structure and supporting tissue-type classification. Beyond visualization, DMT-ME enhances explainability by explicitly linking tissue clusters with relevant gene markers.

To rigorously validate this biological relevance, we performed a quantitative enrichment analysis using ground-truth marker genes from the HCL dataset. As visualized in the expert-gene heatmap in Fig.~\ref{fig_hcl_case_study}(b), the model spontaneously identifies biologically coherent gene modules without supervision. For instance, the \textit{Muscle Module} (including \textit{ACTA2}, \textit{MYH11}, and \textit{TAGLN}~\cite{owens2004molecular, litvinukova2020cells}) is exclusively activated by the musculoskeletal experts (Exp3--5), while the \textit{Neural Module} (characterized by \textit{MBP}, \textit{GFAP}, and \textit{NEFL}~\cite{middeldorp2011gfap, siletti2023transcriptomic}) is distinctly targeted by the nervous system experts (Exp6--7). This precise correspondence between the learned expert selection and known domain knowledge demonstrates that DMT-ME captures meaningful biological signals grounded in the underlying cellular heterogeneity.

\textbf{Potential of Unsupervised Target Discovery.} DMT-ME also shows promise in unsupervised biomarker discovery. Without relying on labels, the model identifies biologically significant genes by associating them with corresponding tissue clusters through expert pathways, as illustrated in Fig.~\ref{fig_hcl_case_study}(b). This capability is useful for exploratory biological studies that aim to identify key genes or biomarkers.

\subsection{Threats to Validity}
\label{sec:threats_validity}

We transparently discuss the limitations of our proposed method. First, regarding \textbf{hardware dependency}, we acknowledge that GPU acceleration provides DMT-ME with computational advantages. To ensure fairness, we report both CPU-only and GPU-accelerated runtimes for all comparable methods (see Section~\ref{sec_exp_Time}). While DMT-ME is optimized for parallel processing, its CPU performance remains competitive, though significantly slower than the GPU version. Second, concerning \textbf{class imbalance}, tests on an artificially imbalanced MNIST (100:1 ratio) showed a performance degradation from 97.0\% to 89.3\%. Future work could integrate weighted sampling or focal loss to mitigate this. Finally, regarding \textbf{generalization}, while hyperbolic embeddings excel on hierarchical structures (e.g., Tree-L, HCL), they provide marginal gains on flat cluster data (e.g., MNIST) compared to Euclidean geometry, confirming that our geometric choice specifically benefits hierarchical topologies.

\section{Conclusion}
This paper proposes DMT-ME, a novel MoE-based explainable deep manifold transformation method for DR, which addresses the limitations of traditional methods by integrating sparse expert specialization with a geometry-aware mapper. Experimental results on image and biological datasets demonstrated that DMT-ME more effectively captures local and global features compared with baseline methods. In addition, the model enhances interpretability through expert-based decision analysis and advanced visualization strategies. 
% Moving forward, we plan to expand DMT-ME's applicability to larger and more diverse datasets, optimizing its computational efficiency for broader real-world applications.

% or


% \appendices

% % you can choose not to have a title for an appendix
% % if you want by leaving the argument blank
% \section{}

\section*{Acknowledgment}
This work was supported by the National Science and Technology Major Project of China~(No. 2021YFA1301603), the National Key R\&D Program of China~(No.2022ZD0115100), the National Natural Science Foundation of China Project~(No. U21A20427), and Project~(No. WU2022A009) from the Center of Synthetic Biology and Integrated Bioengineering of Westlake University. This work was supported by the “Pioneer” and “Leading Goose” R\&D Program of Zhejiang (2024C01140). This work was supported by Key Research and Development Program of Hangzhou (2023SZD0073). We thank the Westlake University HPC Center for providing computational resources. This work was supported by InnoHK program. We also thank the Funding from the ROOTCLOUD TECHNOLOGY CO.,LTD.

\ifCLASSOPTIONcaptionsoff
  \newpage
\fi

\bibliographystyle{IEEEtran}

\bibliography{dimvis,template,ms} 

\end{document}